\documentclass{article}
\usepackage{arxiv}

\usepackage[utf8]{inputenc} % allow utf-8 input
\usepackage[T1]{fontenc}    % use 8-bit T1 fonts
\usepackage{hyperref}       % hyperlinks
\usepackage{url}            % simple URL typesetting
\usepackage{booktabs}       % professional-quality tables
\usepackage{amsfonts}       % blackboard math symbols
\usepackage{nicefrac}       % compact symbols for 1/2, etc.
\usepackage{microtype}      % microtypography
\usepackage{lipsum}
\usepackage{subcaption}
\usepackage{amsmath}
\usepackage{siunitx}
\usepackage{graphicx}
\title{A Spiking Central Pattern Generator for the control of a simulated lamprey robot running on SpiNNaker and Loihi neuromorphic boards}

\author{
  Emmanouil~Angelidis\thanks{Also Chair of Robotics, Artificial Intelligence and Embedded Systems, Technical University of Munich} \\
  Department of Neuromorphic Computing \\  
  fortiss - Research Institute of the Free State of Bavaria \\
  Munich, Germany \\%/ Chair of Robotics, Artificial Intelligence and Embedded Systems, Department of Informatics, Technical University of Munich, Munich, Germany \\
  \texttt{angelidis@fortiss.org} \\
   \And
 Emanuel~Buchholz~\thanks{Also Chair of Robotics, Artificial Intelligence and Embedded Systems, Technical University of Munich} \\
  Department of Neuromorphic Computing \\  
  fortiss - Research Institute of the Free State of Bavaria \\
  Munich, Germany \\%/ Chair of Robotics, Artificial Intelligence and Embedded Systems, Department of Informatics, Technical University of Munich, Munich, Germany \\
  \And
   Jonathan~Patrick~Arreguit~O'Neil \\
  Biorobotics Laboratory, Institute of Bioengineering\\ 
  École Polytechnique Fédérale de Lausanne\\
  Lausanne, Switzerland \\
  \And
  Alexis~Roug\'e\ \\
  Department of Neuromorphic Computing \\  
  fortiss - Research Institute of the Free State of Bavaria \\
  Munich, Germany \\
  \And
  Terrence~Stewart \\
  Computational Neuroscience Research Group\\ 
  University of Waterloo Centre for Theoretical Neuroscience\\
  Waterloo, Canada\\
  \And
  Axel~von~Arnim \\
  Department of Neuromorphic Computing \\  
  fortiss - Research Institute of the Free State of Bavaria \\
  Munich, Germany \\
  \And
  Alois~Knoll \\
  Chair of Robotics, Artificial Intelligence and Embedded Systems\\ 
  Technical University of Munich\\
  Munich, Germany \\
  \And
  Auke~Ijspeert \\
  Biorobotics Laboratory, Institute of Bioengineering\\ 
  École Polytechnique Fédérale de Lausanne\\
  Lausanne, Switzerland \\
}

\begin{document}
\maketitle

\begin{abstract}
Central Pattern Generators (CPGs) models have been long used to investigate both the neural mechanisms that underlie animal locomotion as well as a tool for robotic research. In this work we propose a spiking CPG neural network and its implementation on neuromorphic hardware as a means to control a simulated lamprey model. To construct our CPG model, we employ the naturally emerging dynamical systems that arise through the use of recurrent neural populations in the Neural Engineering Framework (NEF).  We define the mathematical formulation behind our model, which consists of a system of coupled abstract oscillators modulated by high-level signals, capable of producing a variety of output gaits. We show that with this mathematical formulation of the Central Pattern Generator model, the model can be turned into a Spiking Neural Network (SNN) that can be easily simulated with Nengo, an SNN simulator. The spiking CPG model is then used to produce the swimming gaits of a simulated lamprey robot model in various scenarios. We show that by modifying the input to the network, which can be provided by sensory information, the robot can be controlled dynamically in direction and pace. The proposed methodology can be generalized to other types of CPGs suitable for both engineering applications and scientific research. We test our system on two neuromorphic platforms, SpiNNaker and Loihi. Finally, we show that this category of spiking algorithms shows a promising potential to exploit the theoretical advantages of neuromorphic hardware in terms of energy efficiency and computational speed.
\end{abstract}

% keywords can be removed
\keywords{Neurorobotics, Central Pattern Generators, Spiking Neural Networks, Nengo, Neurorobotics Platform, SpiNNaker, Intel Loihi, Neuromorphic Computing, Robotic Control, Hopf Oscillators}

\section{Introduction}
Our work can be placed in the emerging field of Neurorobotics, a field that combines knowledge acquired from different scientific fields and applies them to the study and the control of animal models and robots. Within the context of Neurorobotics, an artificial brain, either biologically or AI inspired, is interacting with a robot model in physical or virtual experiments \cite{knoll_neurorobotics2016}. This enables the testing of hypotheses on virtual embodiment, a concept which encompasses the idea that a brain is not a system isolated from the outer world, but one that constantly receives and processes stimuli and acts according to them.
Neurorobotics problems can fall into various categories, for example robotic control based on cerebellar models \cite{capolei_biomimetic_2019,garrido_alcazar_distributed_2013}, dynamic vision systems based on event-based cameras \cite{kaiser2019embodied2,kaiser2019embodied}, visual perception \cite{bornet_running_2019}, motor control and locomotion tasks \cite{ijspeert_swimming_2007,Bing_2017} and action selection \cite{prescott_robot_2006}. 
\par
A major limitation of existing neuronal models that are often used as artificial brains is that they are both energy and computationally demanding, since they are usually running on conventional CPUs. Even though spiking neural network (SNN) models are computationally sparse by definition \cite{maass_networks_1997}, this characteristic is not taken into account when running them on conventional hardware. Thus specialized hardware that is optimized to run these models has been researched and developed, among others Intel Loihi \cite{Davies_2018}, IBM TrueNorth \cite{Akopyan2015}, SpiNNaker \cite{SpiNNaker} and BrainScale \cite{Schemmel_2010}, the latter two developed within the context of the Human Brain Project. Our work makes use of a SpiNNaker and a Loihi chip that runs the spiking neural network that we developed. 

Many fields of robotics have taken inspiration from biological systems, and particularly from the locomotor system. Locomotion of animals is hypothesized to be controlled to a large extent by functional units in the central nervous system (CNS) called  called Central Pattern Generators (CPGs) \cite{arena_cpg,ijspeert_central_2008}, which are usually described as neuronal systems that create rhythmic activity patterns with minimal sensory feedback. In vertebrates, these locomotor circuits are located mostly in the spinal cord, and receive stimulation from the brainstem and other areas of the brain such as the motor cortex, the cerebellum and the basal ganglia \cite{grillner_biological_2006}. One interesting finding is that these networks are capable of producing rhythmic output in the absence of feedback with minimal stimulation, even if the spinal cord has been completely isolated from the body \cite{grillner_motor_2003}. The investigation of CPG based locomotion control is motivated by the insight that it can give on animals locomotion systems and by the fact that these kind of bio-inspired controllers present good capabilities in terms of autonomy and modulation \cite{survey_cpg_yu}. So far the CPG approach has been largely validated for the locomotion of snake-like robots \cite{crespi_amphibot_nodate, inoue_neural_2004, donati_novel_2016, wang_cpg-inspired_2017}. On an implementation level there exist several CPG models which are formulated as SNNs, and and these spiking CPGs (SCPGs) are often running on specialized or generic neuromorphic hardware. It was shown that such SCPGs running on Neuromorphic hardware such as FPGAs, SpiNNaker or VLSI are providing a robust and efficient way to control a complex movement \cite{cuevas-arteaga_spinnaker_2017} including sensory feedback, namely for bipedal walking \cite{russel_scpg, lewis_cpg_design}, for the movement of an arm \cite{bouganis_training_2010, menon_controlling_2014} or to control a six-legged robot \cite{espinal_design_2016, gutierrez-galan_neuropod:_2019}.

The mathematical modelling of CPGs can be categorized into roughly 3 approaches. The first treats the neural circuitry to the abstraction level of biophysical models and incorporates information about ion pumps and ion channels located in the neural cells membrane and their influence
on membrane potentials and the generation of action potentials, frequently modelled by Hodgkin-Huxley neuron models. The second approach uses simpler leaky integrate-and-fire neurons as the basis of computation, abstracting away low-level biological information. The third category which is also our starting point is deprived of lower level biological information and treats CPGs as systems of nonlinear coupled oscillators, where one oscillator models the activity of a whole oscillatory neural network at an abstract level. Although conceptually the latter is a phenomenological approach based on the observation of the emerging locomotor patterns, it still offers many explanations of the underlying mechanisms of rhythmic pattern generation. One of the first successful attempts to use a high-level mathematical formulation of a CPG and model it as a dynamical system which can be simulated with spiking neurons was the work of Eliasmith and Anderson \cite{eliasmith_rethinking_2000}. Many of the described models are accompanied with neuromechanical simulations that close the loop between body and brain. For an extensive review on CPGs in robotics and biology we refer to \cite{ijspeert_central_2008}.

In this article, we present a high-level SCPG for a lamprey robot that was trained to replicate the dynamics of a system of coupled Hopf-like oscillators. This model is able to produce a set of travelling waves with high-level modulation which correspond to a continuous space of swimming gaits. It can run directly on the neuromorphic SpiNNaker and Loihi boards. It builds on the core Neurorobotics idea of interaction between a virtual robot or animal agent and a virtual brain that runs on neuromorphic hardware and achieves a complex locomotion task. In Section \ref{section:materials_and_methods}, we present the underlying mathematical formulation of the system of coupled Hopf-like oscillators as a first step of the modeling, in Section \ref{section:SNN} we present the spiking version of the CPG and its performance on the two boards. We provide simulations of both the isolated spiking CPG model as well as neuromechanical simulations under different scenarios in \ref{section:results}. We then present our future work (\ref{section:future}) and a conclusion (\ref{section:conclusion}).

\section{Materials and Methods}\label{section:materials_and_methods}
\subsection{Overall model architecture}\label{part:overallArchitecture}
\begin{figure}[b]
\centering
\includegraphics[width=1.0\textwidth]{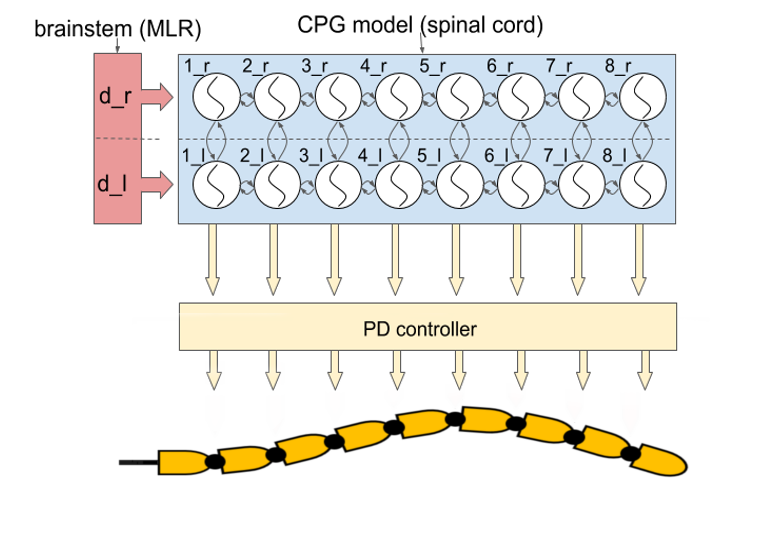}
\caption{The control framework. The brainstem component is abstracting the brain areas that are stimulating the spinal cord, separated into two stimulations, one for each side of the spinal cord. The CPG component, comprised of coupled oscillatory centers organised in a double chain, produces the swimming gaits modulated by the high-level brainstem control. A PD controller is receiving the output of the CPG network and applies it to the robot, controlling the angular rotation of each joint.}
\label{fig:cpg}
\end{figure}

Locomotor CPGs are modulated by higher level control centers of the brain with low-dimensional control signals, a property which makes CPG models good candidates for robotic control problems. This property of CPGs gives them a role similar to a feed-forward controller inside a control framework, of producing oscillatory signals that are modulated by external stimulation. To test whether our CPG model can successfully control a lamprey robot we implemented a neuromechanical simulation for which we employed an accurate 3D model of a lamprey robot that is composed of nine body parts similar to the Amphibot robot in \cite{crespi_amphibot_2005}. These parts are bound together by eight joints that have one degree of freedom: the rotation around the vertical axis. To produce the swimming patterns, the angular positions of these joints oscillate with amplitudes, frequencies and phases prescribed by the CPG model. The complete controller architecture can then be divided in three components (see Figure \ref{fig:cpg}):
\begin{enumerate}
    \item the mesencephalic locomotor region (MLR), that emits high level signals on each side of the spinal cord: the drives;
    \item the central pattern generator (CPG), that generates travelling waves for each joint corresponding to the received drives;
    \item the proportional derivative (PD) controller, that controls the torques applied to the joints to reach the time-varying target angle positions. .
\end{enumerate}

\subsection{Oscillatory signals generation based on coupled abstract Hopf-like oscillators}
In order to explain the synchronization phenomena between the different oscillatory centers in the vertebrate spinal cord, Ijspeert \cite{ijspeert_swimming_2007} proposed a model of nonlinear coupled oscillators, and used this model to control a salamander robot. This model proposes a coupling between different oscillatory centers based on coupling weights that dictate the phase difference and frequency of the oscillatory centers. The oscillators can be chained either in a single or double chain. In the double chain model, the one that we employ here, the activity of the one side of the spinal cord is in antiphase with the activity of the other side, a phenomenon which is also observed in measurements of muscle activity of lampreys. Providing different stimuli, coming from the high-level control centers, between the oscillators found on each side can lead to a shift of the overall oscillatory patterns, which when applied to a robot model induces turning due to the change of the overall curvature of the robot. This dynamical system can be described by the following differential equations which describe a system of phase oscillators with controlled amplitude. The oscillators are described first in phase space, which gives an intuition of how the coupling is induced, and then rewritten in Cartesian space which as we explain is a form suitable for modelling with an SNN:

\begin{equation}  \label{equ:phase}
    \dot\theta_i = 2\pi\nu_i + \sum_jr_jw_{i,j}\sin\left(\theta_i-\theta_j - \Phi_{i,j}\right)
\end{equation}
\begin{equation} \label{equ:amplitude}
    \ddot{r_i} = a_i\left(\frac{a_i}{4}\left(R_i-r_i\right) - \dot{r_i}\right)
\end{equation}
\begin{equation}
    x_i = r_i\left(1 + \cos\theta_i\right)
\end{equation}
\begin{equation} \label{equ:Psi}
    \Psi_i = \alpha\left(x_{i, right} - x_{i, left}\right)
\end{equation}

In this system the $\theta_i$, $v_i$  are the phase and the preferred frequency of the $i$-th oscillator, $r_i$, the amplitude, $x_i$ is the output of the $i$-th oscillator which represents motoneuron activity, and $\Psi_i$ is the output of the model that is applied to the robot and combines the activity of the oscillators of left and the right side of the double chained model. From equation \ref{equ:phase} one can observe that the first derivative with respect to time of the phase of each oscillator, is modulated by the coupling weights $w_{ij}$ and the amplitude of the oscillators it is connected to. It is interesting to note that when the phase differences $\Phi_{ij}$ are reached between the coupled oscillators the term $\theta_j$ - $\theta_i$ - $\Phi_{ij}$ becomes zero, and thus the oscillator oscillates with the preferred frequency $2\pi\nu_i$. This is indeed the case when the steady state is reached, which takes place when certain convergence criteria are met. Equation \ref{equ:amplitude} describes how the amplitude of each oscillator converges to the preferred amplitude $R_i$, with parameter $a_i$ dictating the speed of convergence. This ensures smooth transitions of the amplitude when abrupt changes of the high-level drive occur. Even though this system fully describes a CPG in phase space, it is not suitable for approximation with an SNN, as integrating equation \ref{equ:phase} in time, leads to a constantly increasing phase. This constantly increasing value quickly saturates the representational capabilities of neural populations, as they excel in approximating values within a subset of a larger space. The solution for this problem is to reformulate the problem in Cartesian space as follows \cite{bicanski_decoding_2013}:

\begin{equation}  \label{equ:hopf_x}
    \dot{x_i} = a({R_i}^2 - {r_i}^2)x_i - \overline{ \omega}_i y_i 
\end{equation}

\begin{equation}  \label{equ:hopf_y}
    \dot{y_i} = a({R_i}^2 - {r_i}^2)y_i + \overline{ \omega}_i x_i 
\end{equation}

\begin{equation}  \label{equ:hopf_omega}
    \overline{ \omega}_i = \omega_i + \sum_j\frac{w_{ij}}{r_{i}}[(x_iy_j-x_jy_i)\cos{\Phi_{i,j}} - (x_ix_j-y_iy_j)\sin{\Phi_{i,j}}]
\end{equation}

where $x_i$, $y_i$ denote the x and y-coordinates of a point in 2-D space moving in a circle through time, with frequency controlled by equation \ref{equ:hopf_omega}. The parameter $a$ dictates the speed of convergence of the amplitude to the steady state, and $r_i$ it the norm of the [x,y] vector. This formulation is close to the standard form of coupled Hopf oscillators with coupling to other oscillators. This equation has the advantage that the x,y values stay within a limit cycle, whose radius is dictated by the amplitude of the oscillation, solving the problem of continuously increasing phase when one attempts to use the phase representation.

To incorporate the drive corresponding to the high-level stimulation we use two piece-wise linear functions, which saturate when the stimulation is outside of a certain range. These two functions control the target frequency and the target amplitude of each oscillator according to the relations:

\begin{equation} \label{high_level_omega_drive}
    \omega_i(d)= 
\begin{cases}
    c_{\omega,1}d + c_{\omega,0},& \text{if } d_{low}\leq d \leq d_{high}\\
    0, & \text{otherwise}
\end{cases}
\end{equation} 

\begin{equation} \label{high_level_R_drive} 
    R_i(d)= 
\begin{cases}
    c_{R,1}d + c_{R,0},& \text{if } d_{low}\leq d \leq d_{high}\\
    0, & \text{otherwise}
\end{cases}
\end{equation} 

These two equations replicate biological observations that the frequency and amplitude of muscle contraction increase together with increased stimulation, hence leading to faster locomotion.
They complement the CPG with high-level modulation, and with them we have a complete mathematical formulation of the control framework, which we implement in an SNN.

\subsection{Implementation of the coupled oscillators system in a spiking network }\label{section:SNN}

\subsubsection{Architecture of the spiking CPG neural network} \label{architecture}
\begin{figure}[ht!]
\centering
\includegraphics[width=1.\textwidth]{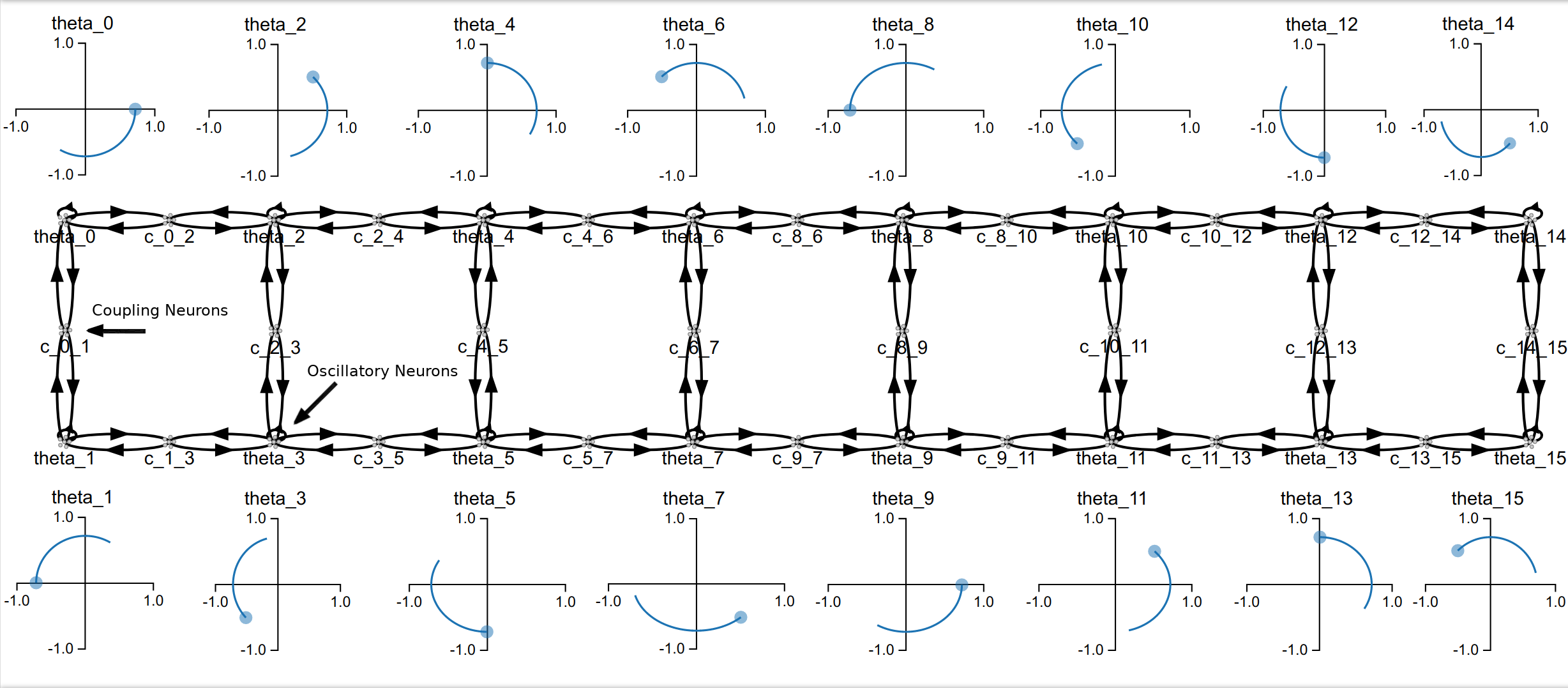}
\caption{Architecture of the spiking CPG model. Each oscillatory center, noted with $theta_i$ is coupled with its neighbours through an intermediate population, depicted with $C_{ij}$. The intermediate population is computing the coupling term of equation \ref{equ:hopf_omega}. The x-y diagrams corresponding to each oscillator show the trajectory of a point traversing the limit circle through time for the ideal mathematical model. As can be observed, the oscillators in each side of the spinal cord have an antiphase relationship between them, whereas the ones upwards or downwards have a fixed phase difference of $4\pi / NumOsc$.}
\label{fig:nengo_double_chain_architecture}
\end{figure}

The model that we introduced in the previous section is a mathematical formulation of a system of coupled abstract Hopf-like oscillators, modulated in frequency and amplitude by high-level stimulation. We show that such a system can be easily simulated with an SNN simulator. To do so we designed a modular SNN architecture where one oscillatory center is represented by one population of spiking neurons and computes the equations described in (\ref{equ:hopf_x} - \ref{equ:hopf_omega}). This population at the same time encodes equation \ref{high_level_R_drive}. For the coupling between the neural oscillators we introduce an intermediate population which receives the x,y values from neighbor oscillators, and computes the coupling term of equation \ref{equ:hopf_omega}. This intermediate population facilitates the exchange of data between the neural oscillators, and it's presence is dictated purely by the framework that we chose to implement the SNN. The overall architecture of the model can be seen in Figure \ref{fig:nengo_double_chain_architecture}. At the same time each of the oscillatory centers is receiving input from the high-level drive through equations \ref{high_level_omega_drive} - \ref{high_level_R_drive}. 
\subsubsection{Choice of the neural simulator}
In order to replicate the system of modulated oscillators with a spiking neural network the choice of a framework that can perform such numerical computations was necessary. A characteristic shared by most neural simulators is that they allow the simulation of simple leaky integrate-and-fire neuron models (LIF). According to this model  \cite{Gerstner:2002:SNM:583784} the neuron spikes when its membrane potential reaches a certain threshold. Each neuron is excited by the neurons that are connected to it either in an excitatory or inhibitory fashion, increasing or decreasing the membrane potential respectively. After a period of inactivity the membrane potential is reset -leaks- to a base value. A neuron is usually connected with multiple other neurons via junctions called synapses. The information flow from one neuron to the other is dictated among other factors by the level of present in the synapse neurotransmitters and whose release is regulated by dedicated proteins. The overall strength of the connection between neurons is dictated by the synaptic weight. From a computational perspective, the adaptation of the synaptic weights through synaptic plasticity mechanisms is the process which allows these networks of neurons to learn a representation. Synaptic plasticity mechanisms can be either biologically accurate, i.e. STDP \cite{markram_spike-timing-dependent_2012}, or variations of some machine learning inspired approach such as the ones making use of backpropagation algorithms \cite{bellec_long_2018}, or biologically plausible mechanisms such as the e-prop algorithm \cite{bellec_biologically_2019}. Most computational models of spiking neurons employ the simple Leaky integrate-and-fire neuron model. We use these types of neurons for our study as well.  Several simulation platforms were suitable for the task of simulating such neurons, but Nengo \cite{bekolay_nengo:_2014} was chosen for two reasons.  First, it has built-in methods for generating neural networks that approximate differential equations.  This approach is described in section \ref{Nengo_framework}.  Second, it can generate versions of these networks that can run on dedicated neuromorphic hardware, as we discuss in section \ref{section:hardware}.

\subsubsection{Nengo and the Neural Engineering Framework}
\label{Nengo_framework}
In this section we give an overview of the Neural Engineering Framework (NEF), which is a general methodology for creating neural networks that approximate differential equations \cite{eliasmith_neural_2003}.  Importantly, it generalizes to any neuron model, including LIF spiking neurons, and takes into account the timing of synapses.

To understand the NEF, we start with the standard observation that a normal feed-forward neural network is a function approximator.  That is, if we have some input $x$ and some output $y$, we can train a neural network produce the desired output $y=f(x)$.  While this training can be done using any neural network learning algorithm, here we just use the simple method of having a network with a single hidden layer of LIF neurons (no non-linearities at the input or output), randomly generate the first layer of weights, and use least-squares minimization to solve for the second layer of weights.  This method works for a large range of functions and is robust to spiking neuron models \cite{eliasmith_neural_2003}.

However, to generate the CPG model described here, we need networks that approximate differential equations.  Here, the NEF applies the following method.  Suppose we want the differential equation $\dot x = f(x, u)$.  We build a feed-forward network where the inputs are $x$ and $u$ and the output approximates $\tau f(x, u) + x$.  We introduce the variable $\tau$ here, which will be used as the time constant of a simple exponential low-pass filter synapse that will connect the neurons.  Now to generate the recurrent neural network, we simply connect the output of that network back to itself, and scale the $u$ input by $\tau$. The resulting network will approximate $\dot x = f(x, u)$.  See \cite{eliasmith_neural_2003} for a full proof, which is based on the observation that the Laplace transform of the low-pass filter is $F(s)=1/(1+s\tau)$.  Similar transformations can be done for more complex synaptic filters, but we do not use those here.

\begin{figure}[ht!]
\centering
\includegraphics[width=1\textwidth]{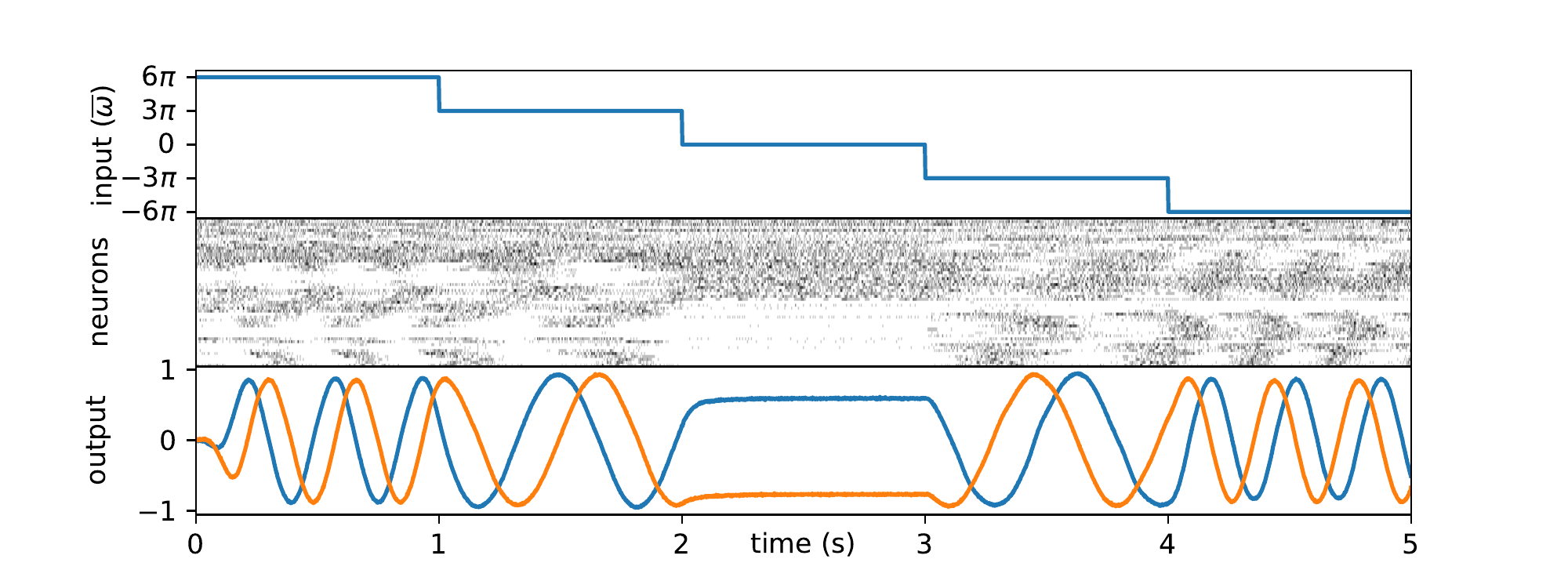}
\caption{The behavior of a single Hopf-like oscillator implemented in spiking neurons using Nengo and the Neural Engineering Framework (NEF).  The model consists of an all-to-all recurrently connected layer of LIF neurons with exponential synapses with 100ms time constants.  Their spiking activity is shown in the middle row, sorted by similarity.  A single input ($\overline{\omega}$) is provided, and the two outputs show that it functions as a controlled oscillator.  The input weights,
recurrent weights, and output weights are found using the NEF such that the network approximates $\dot{x} = a({R}^2 - {r}^2)x - \overline{ \omega} y$ and $\dot{y} = a({R}^2 - {r}^2)y + \overline{ \omega} x$.} 
\label{fig:hopf_model}
\end{figure}

As an example of this process, Figure \ref{fig:hopf_model} shows an NEF model of a single Hopf-style oscillator.  This was formed by creating a feed-forward single-hidden-layer neural network with three inputs ($x$, $y$, and $\overline{\omega}$) and two outputs ($\tau(a({R}^2 - {r}^2)x - \overline{ \omega} y)+x$ and $\tau(a({R}^2 - {r}^2)y + \overline{ \omega} x)+y$).  The weights for this network were found by randomly sampling the inputs ($x$, $y$, and $\overline{\omega}$), computing the desired outputs for each input, and then training the network given this data.  Afterwards, the resulting input and output connection weights were multiplied together to create the recurrent neural network shown.

The Nengo software toolkit  \cite{bekolay_nengo:_2014}, which is the software implementation of the more general Neural Engineering Framework, provides high-level tools for creating such networks for a variety of neuron models.  Crucially, it also provides facilities for linking networks together, so that large systems can be built out of these components.  Futhermore, the resulting systems can be automatically compiled to run on CPUs, GPUs, or a variety of neuromorphic hardware.

%[[[TODO Terry: add an example of a basic Hopf oscillator running in Nengo to depict the above process]]]

\subsubsection{The Nengo model}
\label{nengo_model}
\begin{figure}[ht!]
\centering
\includegraphics[width=1.\textwidth]{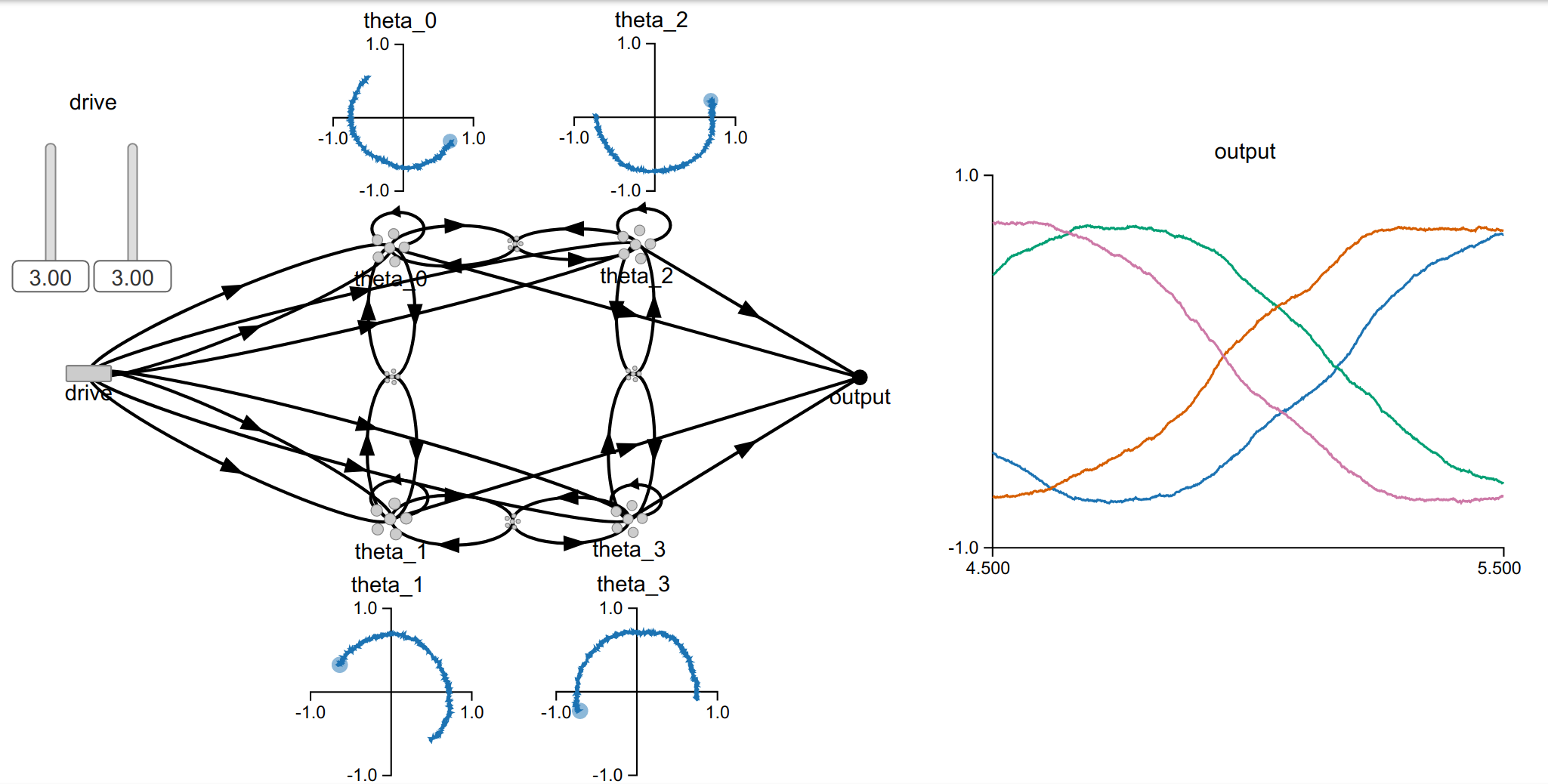}
\caption{(Left) The Nengo simulated model where 4 oscillatory centers are shown. In this simulation the high-level stimulation is driving the oscillations. (Right) The output of each oscillator that corresponds to the decoded spiking activity, when 2000 neurons per oscillatory center are used, is depicted.}
\label{fig:nengo_4_oscillators}
\end{figure}

Based on the third principle of the NEF we employ the dynamical systems that emerge through the use of recursive neurons to implement the oscillators in our model. It is worth noting that recurrent neural populations can implement various dynamical systems, such as integrators, oscillators, even chaotic systems such as Lorenz attractors. The network computes each function from equations(\ref{equ:hopf_x}-\ref{high_level_R_drive}) according to the NEF principles. By doing so the decoded spiking activity of each neural population can be seen as a real-valued vector with the appropriate dimensions. For the populations that encode the oscillators (depicted with $theta_i$ in Figure \ref{fig:nengo_double_chain_architecture}) this 4-dimensional vector represents the values $[\dot x,\dot y, \omega, R]$. For the intermediate neuron populations that compute the coupling part of equation \ref{equ:hopf_omega} the 4-dimensional vector represented is $[\dot{x_i},\dot{y_i},\dot{x_j},\dot{y_j}]$. The high-level drive is approximated by the decoded activity of a neuronal population dedicated in receiving the drive and translating it to neural activity. A dedicated readout output node (non-spiking) can be used to read the decoded output of the system, that corresponds to the x-coordinate of the Hopf-like oscillator. The complete system with input and output for 4 oscillatory centers can be seen in Figure \ref{fig:nengo_4_oscillators}. As will be shown the system can scale to a larger number of oscillatory centers but the scaling can be limited by the capabilities of the neuromorphic hardware that it is running on.

As mentioned in \ref{Nengo_framework} the Neural Engineering Framework can be used to approximate any linear or non-linear function with spiking activity by computing the connection weights between the different components of a spiking neural network, acting as a neural compiler. This alleviates the need for explicit training of the SNN, as in the NEF the information that needs to be provided is limited to the properties of the neurons (i.e. membrane threshold potential, neuron types), the values that the neural populations need to represent and the functions that they compute, and the NEF solves for the connection weights that will compute the desired functions. This enables specifying the high-level mathematical functions that are encoded by the SNN and that works both for feed-forward as well as for recurrent connections. The latter is particularly relevant for our work as it enables dynamical systems such as the oscillator system that we employ to emerge from the neuronal activity. In order for the connection weights to be computed by the NEF, during the initialization phase of the simulation a random selection of sampling points to be used as inputs to the function to approximate is selected. These points are based on the input space that the neuronal population approximates, f.e. points in the space [0,1] for a population that encodes 1-D values. Then these points are used to generate training data from the functions, by providing the points as inputs to the desired functions and collecting the output. Subsequently a least-squares optimization computes the weights that best fit the decoded neuronal activity to the training data. For a more detailed technical overview of this method we refer the viewer to \cite{Stewart2012ATO}.

\subsubsection{Perturbations and robustness of the CPG model}
\label{section:robustness}
Animal CPGs have been documented to adapt to various perturbations (i.e. external application of a force), by reacting smoothly and exhibiting stable limit cycle behavior, i.e. recovering the gait patterns without losing synchronization. Furthermore different degrees of stimulation of the oscillatory centers on the spinal cord can lead to different gaits. Simple asymmetrical stimulation between the right and left side drive of the spinal cord can induce a shift of the gait patterns to the left or to the right, and can induce turning. We show that these characteristics are exhibited by our model under the following scenarios:

\begin{enumerate}
  \item Perturbation of a single oscillatory center by external stimulation
  \item Asymmetrical stimulation of the spinal cord from left to right side of the spinal cord
\end{enumerate}

These scenarios show the CPG model's ability to quickly recover under external perturbations as well as to modulate swimming gaits.

\subsection{Neuromechanical simulation in the Neurorobotics Platform}
To test the output and the high-level adaptation of the control signals we performed a closed-loop neuromechanical simulation of our model with a robot model as a body. The motivation behind simulating our model within a physical simulation framework comes from the fact that neural circuits and control algorithms cannot be separated from their natural habitat, the body. Only within an embodied simulation can we test whether the system that we propose can successfully control a robot. For such a full closed-loop robot-brain interaction simulation we made use of a framework built exactly for this purpose, the Neurorobotics Platform. The Neurorobotics Platform (NRP) is a software simulator developed within the Human Brain Project \cite{falotico_connecting_2017} that enables the synchronization and exchange of data between modelled brains and virtual robots within a physical simulation environment. The Robotic Operating System \cite{Quigley09} is the middleware which enables the communication between the different software components, which is also supported by a multitude of physical robots. Within the NRP there is no need for an explicit synchronization mechanism between the physical world and the modelled brain, as such a mechanism is built into the framework. The physical simulation is provided by Gazebo \cite{Koenig:gazebo}, which interfaces with multiple physics engines. It supports directly many different brain simulators such as NEST \cite{Gewaltig:NEST}, Nengo and SpiNNaker, and through Nengo one can run models on Loihi. We used this framework to connect the Nengo model presented in section \ref{nengo_model} with the lamprey robot (Figure \ref{fig:cpg}).

To complement the simulation with a simplified fluid dynamics model, we implemented a drag model, which is computing the forces produced by the swimming motion, forcing the robot to move forward. The drag model is the one presented in \cite{ekeberg_combined_1993}, and computes the forces applied on each robot link based on the formulas:

\begin{equation}
   E_{i\parallel } = \lambda_{i\parallel }\upsilon_{i\parallel}^2
\end{equation}

\begin{equation}
   E_{i\bot } = \lambda_{i\bot }\upsilon_{i\bot}^2
\end{equation}

and the coefficients $\lambda$ can be computed by

\begin{equation}
   \lambda_{i\parallel } = \frac{1}{2}C_{i\parallel }S_{i}\rho
\end{equation}

\begin{equation}
   \lambda_{i\bot } = \frac{1}{2}C_{i\bot }S_{i}\rho
\end{equation}

where $\upsilon_i\parallel$ and $\upsilon_i\bot$ are the velocity components of each link relative to the water in the parallel and perpendicular directions. The parameter $\lambda$ depends on the fluid density $\rho$ and the parameter $S_i$ is the surface of the link perpendicular to the link movement. This drag model is only a simple approximation of the fluid forces applied on the robot, but offers simplicity and computational speed compared to the 3D Navier-Stokes equations.

\subsubsection{The neuromechanical simulation scenarios} 
\label{subsection:scenarios}
We tested the arising swimming gaits under different simulation scenarios. Firstly we show that the spiking CPG can produce swimming even with a low number of neurons. Secondly we show unperturbed swimming with no high-level modulation. Thirdly, we present modulation of the swimming by the high-level drive with control of direction and speed. To show the ability of the controller to incorporate sensory feedback from the simulation dynamically we add a water speed barrier to the simulation. This speed barrier forces the robot to move to the side without adaptation of the high-level drive, but with modulation the robot manages to overcome it. The water speed barrier is implemented in the form of a global fluid velocity vector opposite to the forward direction. A summary of the scenarios:

\begin{enumerate}
  \item Unperturbed swimming, effect of varying number of neurons per neural population
  \item Unperturbed swimming, no high-level modulation
  \item Unperturbed swimming, control of the speed and direction of the robot
  \item Presence of water speed barrier, no high-level modulation
  \item Presence of water speed barrier, high-level modulation
\end{enumerate}

The method that we used to modulate the high-level drive of the robot in the presence of a speed barrier consists of a high-level feedback loop that modulates the turning commands (i.e. the left-right asymmetry of drive signals) towards a desired target angle (e.g. similarly to a fish aiming to swim towards a particular far away target). This is implemented through a linear minimization of the error between a target global angle around the z-axis of the robot's head and the actual angle of the robot's head around the z-axis. Thus, when the robot turns i.e. left, the error between the target angle and the measured angle increases and the right drive increases linearly to compensate for the deviation from the target angle. The equations that we used for this strategy:

\begin{equation}
    d_{right}=  
\begin{cases}
    d_{right0} + CF*abs(\vec{R}_{z,target} - \vec{R_z}),& \text{if } R_z - R_{z,target} \leq 0 \\
    d_{right0} & \text{otherwise}
\end{cases}
\end{equation}

\begin{equation}
    d_{left}=  
\begin{cases}
    d_{left0} + CF*abs(\vec{R}_{z,target} - \vec{R_z}),& \text{if } R_z - R_{z,target} \geq 0 \\
    d_{left0} & \text{otherwise}
\end{cases}
\end{equation}

Where the left drive is increased when the error is positive, and the right when negative. $\vec{u}_{target}$ is the target lateral velocity, $\vec{R}_{z}$ is the recorded rotation around the z-axis of the robot's head, $CF$ is the correction factor that linearly multiplies the error, and $d_{right0}$ and $d_{left0}$ provide the baseline of the drive stimulation. This simple error correction strategy proves to be enough to correct the deviation of the robot from a target angle by modulating the CPG with the high-level drive.

\subsection{Nengo on SpiNNaker-3 and Loihi boards} \label{section:hardware}
% TODO Terry add some more info on the technical considerations of Nengo Loihi / Spinnaker
As stated in \cite{an_roadmap_2018}, the computational limitations for running spiking models on conventional CPUs are originating in the von Neumann architecture. Conventional computers are built and optimized to perform Boolean algebra operations and arithmetic on the data stored in memory. Hence, this data needs to be transferred back and forth between the memory and the CPUs, which can be time consuming. Neuromorphic hardware on the other hand is specialized in running spiking neural networks. The computation takes place in many small calculators that have access to a small amount of local data. This strategy reveals itself to be more time and energy efficient for neuron oriented computations. For this reason, we tested our Nengo model on a SpiNNaker-3 \cite{SpiNNaker} and a Loihi board \cite{Davies_2018}. Due to the direct connection of SpiNNaker and Loihi boards to Nengo with a software interface our model remained high-level but could be run directly on the boards.

It should also be emphasized that, for efficiency reasons, the actual neuron model running on conventional CPUs, SpiNNaker-3, and Loihi, are all slightly different.  They can all implement Leaky Integrate-and-Fire neurons (and other neuron models), but they all make slightly different approximations (e.g. fixed-point rounding).  This means that the optimal neural network connection weights for these different hardware platforms will all be slightly different.  However, because we specify our model in Nengo using only the mathematical function to be approximated, this means that Nengo can take the hardware details into account when solving for the connection weights, and the user does not have to modify their model to adjust for different hardware platforms.

That said, there are still some areas where the Nengo-SpiNNaker and Nengo-Loihi interfaces have room for improvement.  In particular, the software support for automatically splitting a group of neurons to run across multiple hardware cores is lacking, effectively giving an upper limit on the size of a single group of neurons that is hardware-dependent. We also encountered hardware limitations on the amount of data that could be probed (i.e. recorded) during the running of the simulation, as discussed in Section \ref{section:hardware_metrics}.

\section{Results} \label{section:results}
\subsection{Running the isolated CPG model} \label{section:isolated_cpg_run}
\begin{figure}[ht!]
\centering
\includegraphics[width=1\textwidth]{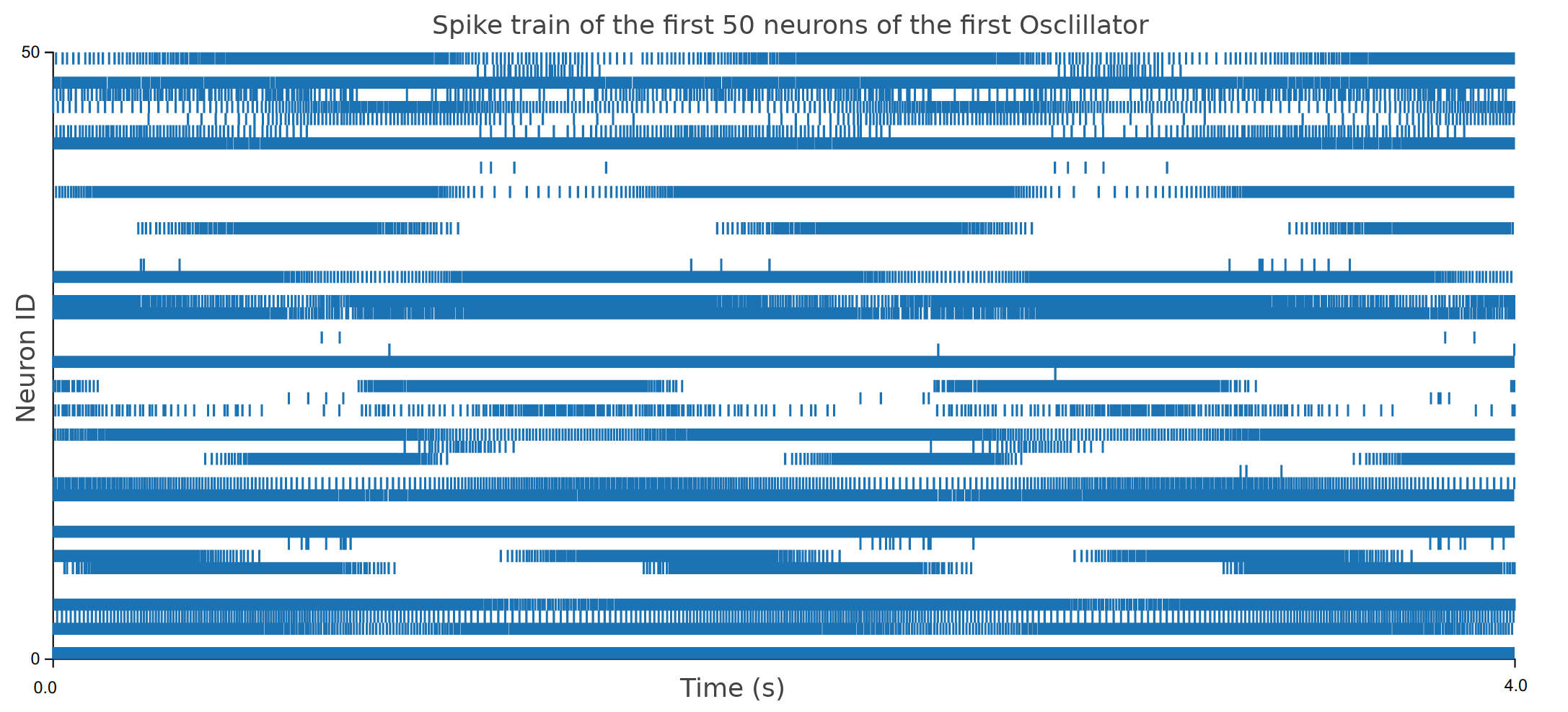}
\caption{Spike train of the first 50 neurons of an oscillatory population with 2000 neurons for 4 secs. The activity of the neurons shows clears signs of periodicity. The neurons are continuously alternating between high and low firing rates.}
\label{fig:spike_train}
\end{figure}

\begin{figure}[ht!]
\centering
\includegraphics[width=1\textwidth]{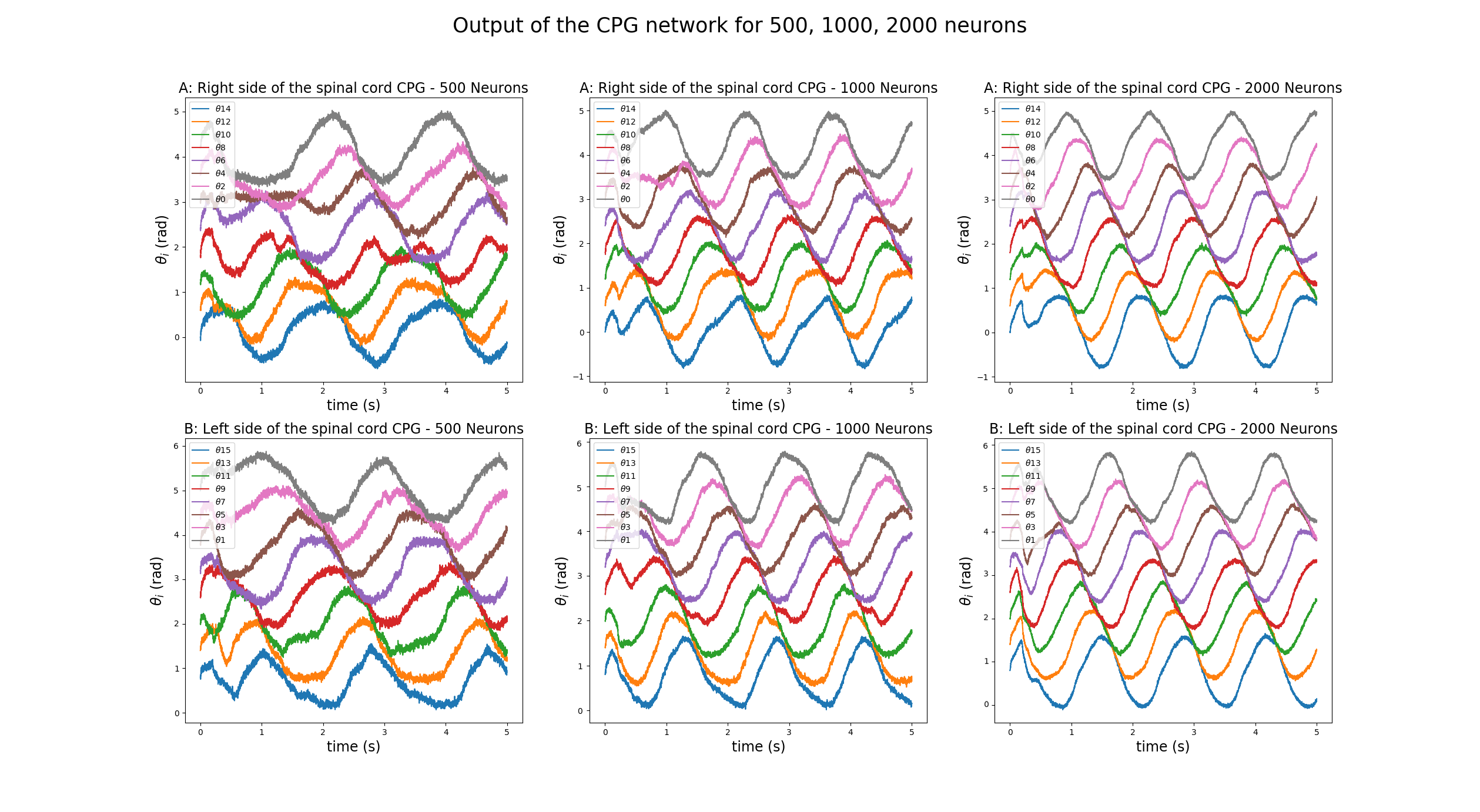}
\caption{The output of the network for different number of neurons per oscillatory population. Even with 500 neurons the network can produce an oscillatory output, of lower quality as some of the oscillators' waves are not smooth and there is more high-frequency noise. With 100 neurons there is an improvement of the quality of the signals, whereas with 2000 neurons the signals are smooth and without high-frequency noise. Even with a low number of neurons the patterns are capable of producing simulated swimming. The network was trained in Nengo with a random seed of 0.}
\label{fig:different_neurons}
\end{figure}

The first test that we performed on the isolated (i.e. no time-varying external modulation) spinal cord model, shows that our system can produce oscillations and traveling waves from random initial conditions meaning that it exhibits limit cycle behavior. For such a scenario there is a clear periodic activation of the spiking neurons inside the oscillatory populations as can be seen in \ref{fig:spike_train}. In order to provide benchmarks for the neuromorphic platforms vs the CPU as well as to show the adaptive capabilities of our model we ran the model with different numbers of neurons and different numbers of oscillatory centers. An interesting finding is that oscillatory patterns are generated even with low numbers of neurons as can be seen in Figure \ref{fig:different_neurons}.

\begin{figure}[ht!]
\centering
\includegraphics[width=1\textwidth]{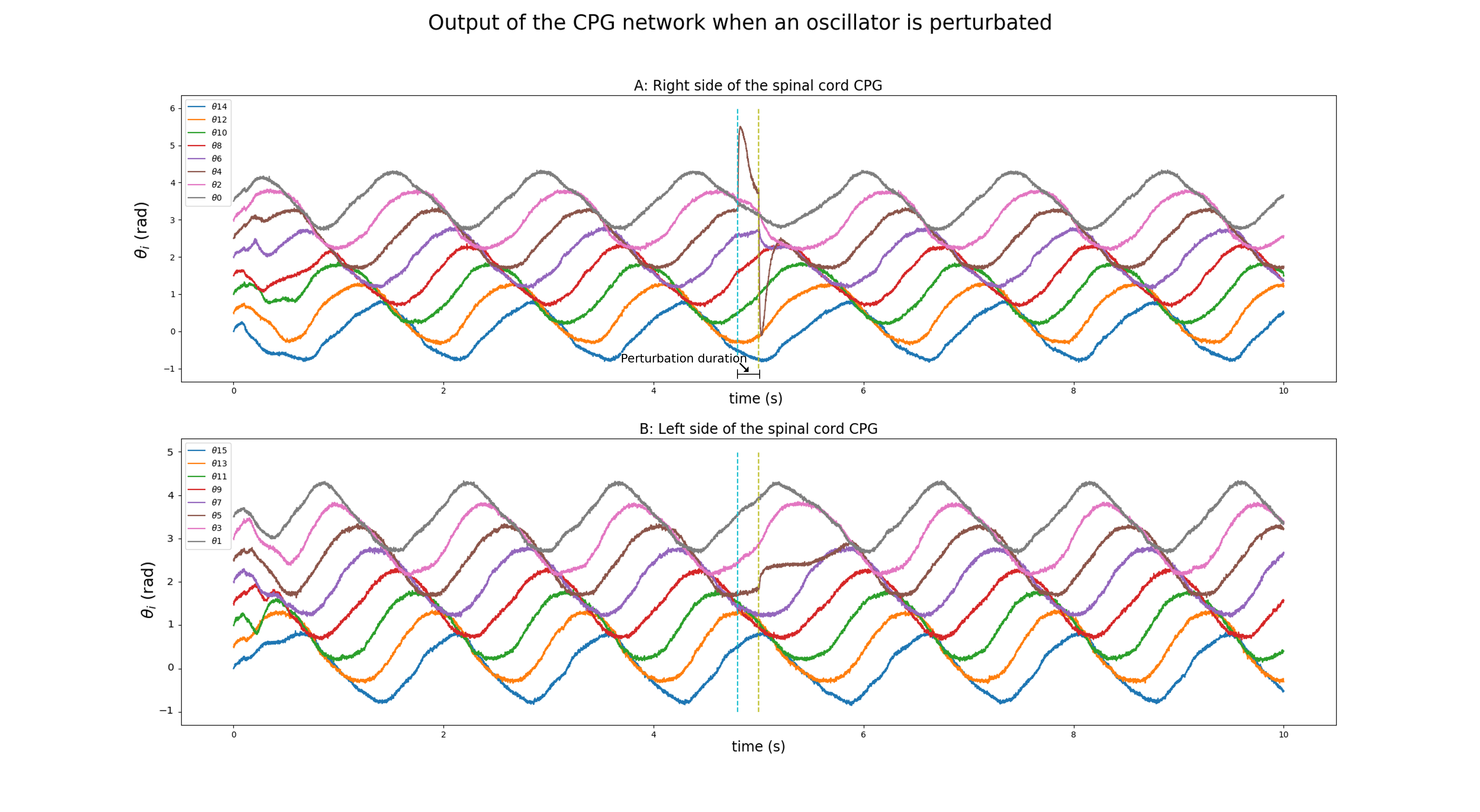}
\caption{The output of the network when the 5th oscillator is perturbed by an external signal. The perturbation lasting from 4.8 to 5 secs causes disturbance of the neighbouring oscillators' $\theta_2$, $\theta_5$, $\theta_6$ wave patterns. The model quickly recovers when the perturbation is removed.}
\label{fig:perturbation_5_th_oscillator}
\end{figure}

Furthermore, perturbing the model by providing explicit stimuli on specific oscillatory centers, can lead to some interesting behaviours which show the stability of the circuit. As can be seen in Figure \ref{fig:perturbation_5_th_oscillator} a single external perturbation on one of the oscillatory centers leads to a temporary disruption of the signals, localized around the neighbouring oscillatory centers. Upon removal of the perturbation the oscillators quickly recover and stabilize. This is the limit cycle property of the high-level mathematical model that is captured well by the spiking network, and exhibits the robustness of the model, a property which is of particular importance for robotics problems.

\begin{figure}[ht!]
\centering
\includegraphics[width=1\textwidth]{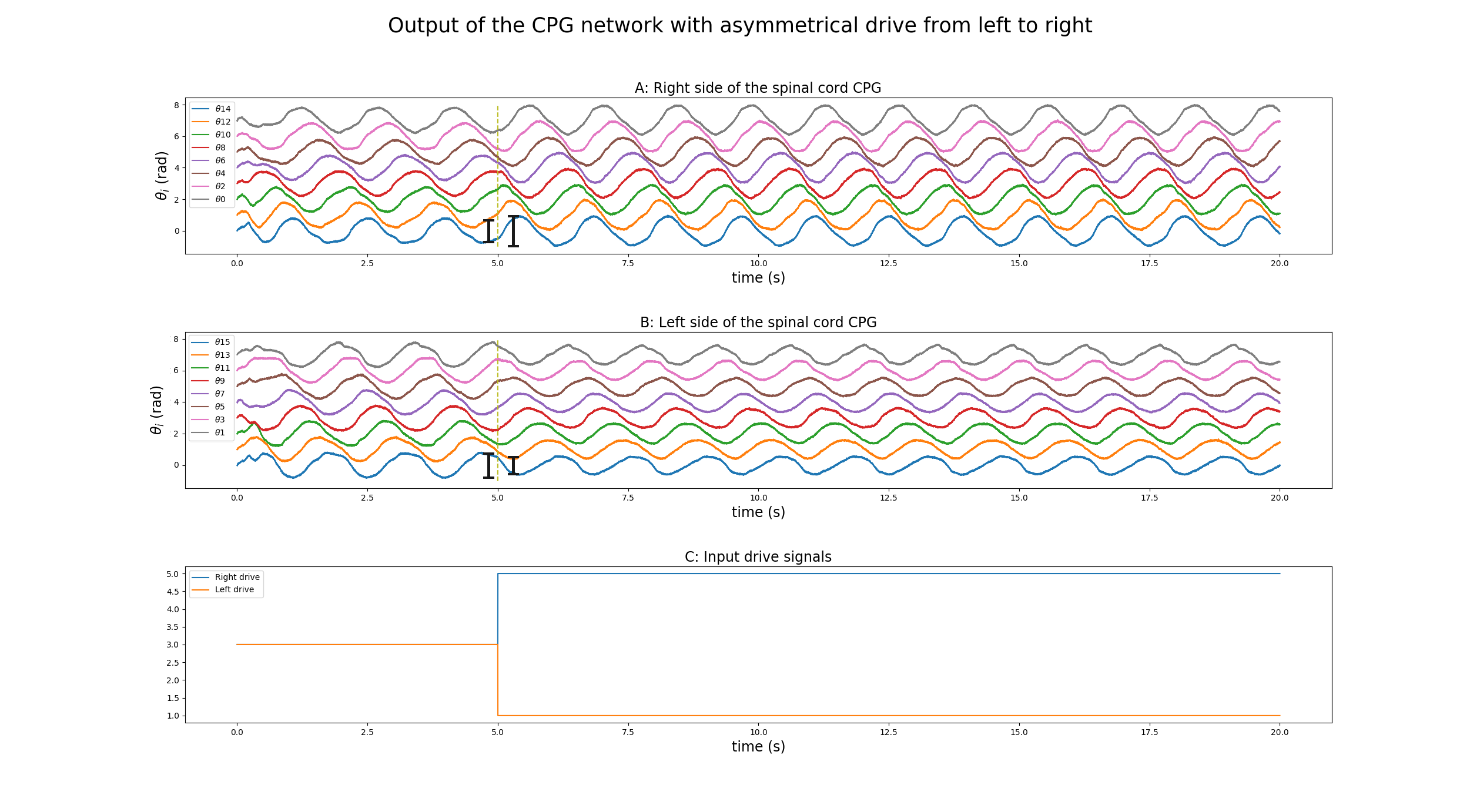}
\caption{The output of the CPG network for 16 oscillatory centers, where each oscillator is depicted with $\theta_i$. An asymmetric drive is provided to the network after 5 seconds of simulation, increasing the drive on the right side of the spinal cord, and decreasing it on the left. As can be observed the amplitude of the oscillations on the right side increases, whereas on the left side decreases.}
\label{fig:asymmetrical_left_right}
\end{figure}

The high-level modulation and control of the signals when varying the input to the network under the scenario described in \ref{section:robustness} can be seen in Figure \ref{fig:asymmetrical_left_right}. In this scenario a simple asymmetrical variation of the input signals between the left and the right side of the spinal cord leads to a formulation of different travelling wave patterns, which can induce different swimming behaviours. A variation between the left and right side of the spinal cord leads according to equation \ref{equ:Psi} to a shift of the center of the signals towards positive or negative angles, which in turn induces a shift of the joints angles towards one side, causing the robot's curvature to change, inducing a change of direction.

\subsection{Neuromechanical simulations}
\subsubsection{Unperturbed swimming}
\begin{figure}[ht!]
\centering
\includegraphics[width=1.0\textwidth]{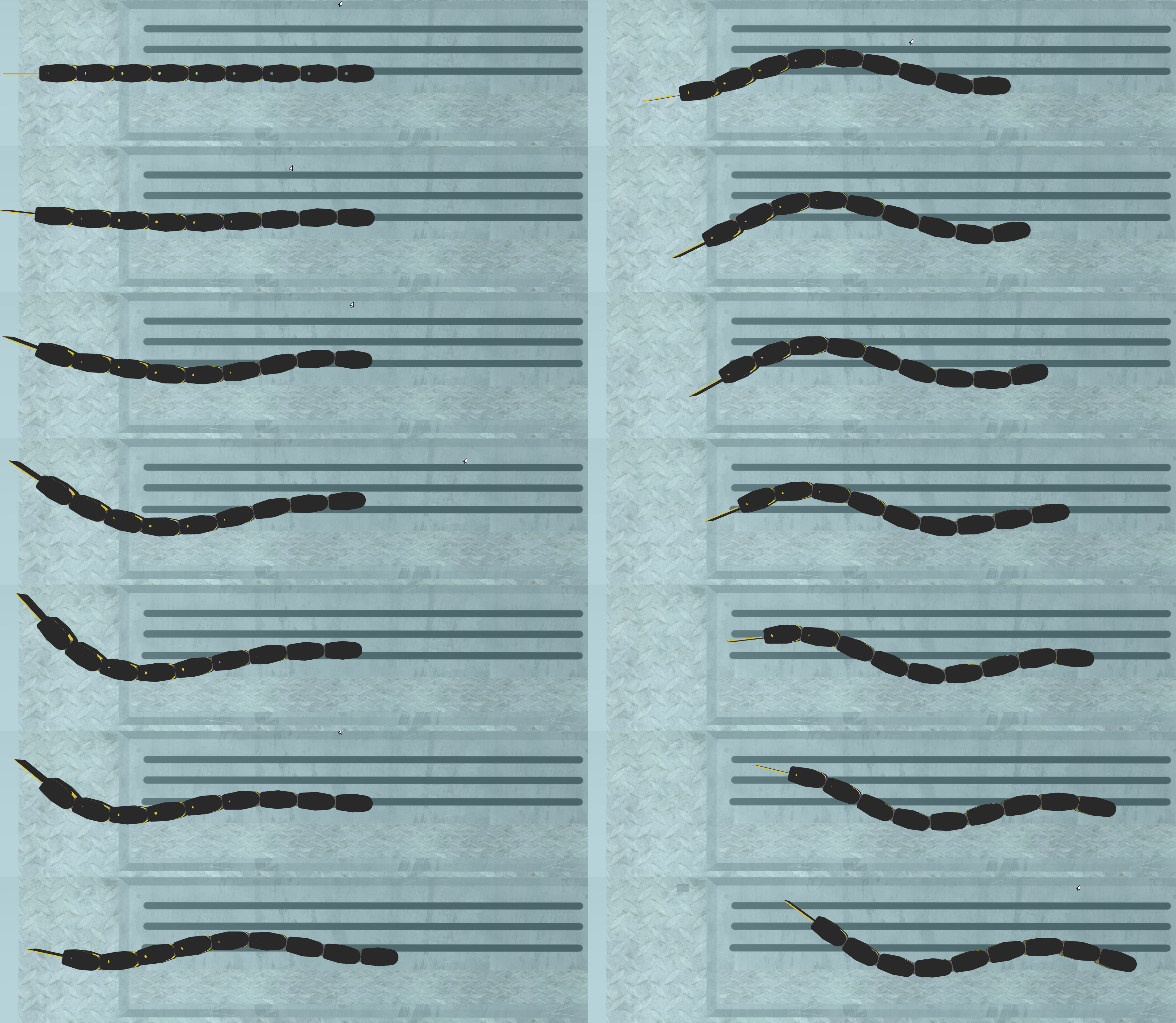}
\caption{Swimming with the simulated robot, with snapshots at 160ms intervals for the unperturbed non-adaptive scenario. The network consists of 2000 neurons per neural population. The travelling wave is propagated along the robot's body from head to tail. }
\label{fig:unperturbed_no_adapt_robot_config}
\end{figure}

As mentioned in section \ref{section:isolated_cpg_run} swimming patterns arise even with a smaller number of neurons for every neural population in the spiking neural network, albeit the fewer neurons the less precise the approximation is. A comparison of the three simulation scenarios with consecutively larger numbers of neurons can be seen in videos \footnote{https://youtu.be/E27Zj1ShI14} (500 neurons), \footnote{https://youtu.be/b1E9EvVRoOw} (1000 neurons), \footnote{https://youtu.be/n3Q-Sn6jUKU} (2000 neurons). The robot configurations in the scenario of the 2000 neurons can be seen in Figure \ref{fig:unperturbed_no_adapt_robot_config}. The videos correspond to Figure \ref{fig:different_neurons}, and as can be observed the less neurons, the less smooth the swimming is. Nevertheless, even the 280 neurons per neural population are enough to provide a swimming pattern. 

Asymmetry of the driving signals between left and right induces turning as can be seen in video \footnote{https://youtu.be/DKdtTFdthbI}, and providing such drives is a simple way to navigate the robot towards one direction. Using a closed loop control method such as the one described in \ref{subsection:scenarios} such asymmetries can be computed and provided automatically to the control loop.
\subsubsection{Presence of water speed barrier} \label{subsecion:speed_barrier}
As described in section \ref{subsection:scenarios}, to demonstrate the controllability of the robot with a closed loop controller we examine the behaviour of the robot with the presence of a speed barrier, first without adaptation of the high-level signal \footnote{https://youtu.be/hCtUVjqVr5g} and then with high-level adaptation \footnote{https://youtu.be/Q58Me79cfSs}. In the first video, the speed barrier causes the robot to follow a trajectory towards the side, by applying higher drag forces to the robot in the lateral direction. In this scenario the robot does not manage to compensate for the presence of the speed barrier as the unmodulated oscillatory signals do not induce a correction of the direction of the robot. In the second video on the other hand, the error correction mechanism described in \ref{subsection:scenarios} is activated, causing the trajectory of the robot to be corrected to compensate for the speed barrier, and eventually it manages to orient itself and swim forward. We can observe that the model adapts well when the high-level tonic drive signal is regulated by the error correction mechanism, which conceptually corresponds to the adaptation that a decision making center of the brain would perform in order to follow a certain trajectory. 

\subsubsection{Energy and computational speed metrics on SpiNNaker-3 and Loihi boards}
\label{section:hardware_metrics}
\begin{figure}[ht!]
\centering
\includegraphics[width=1\textwidth]{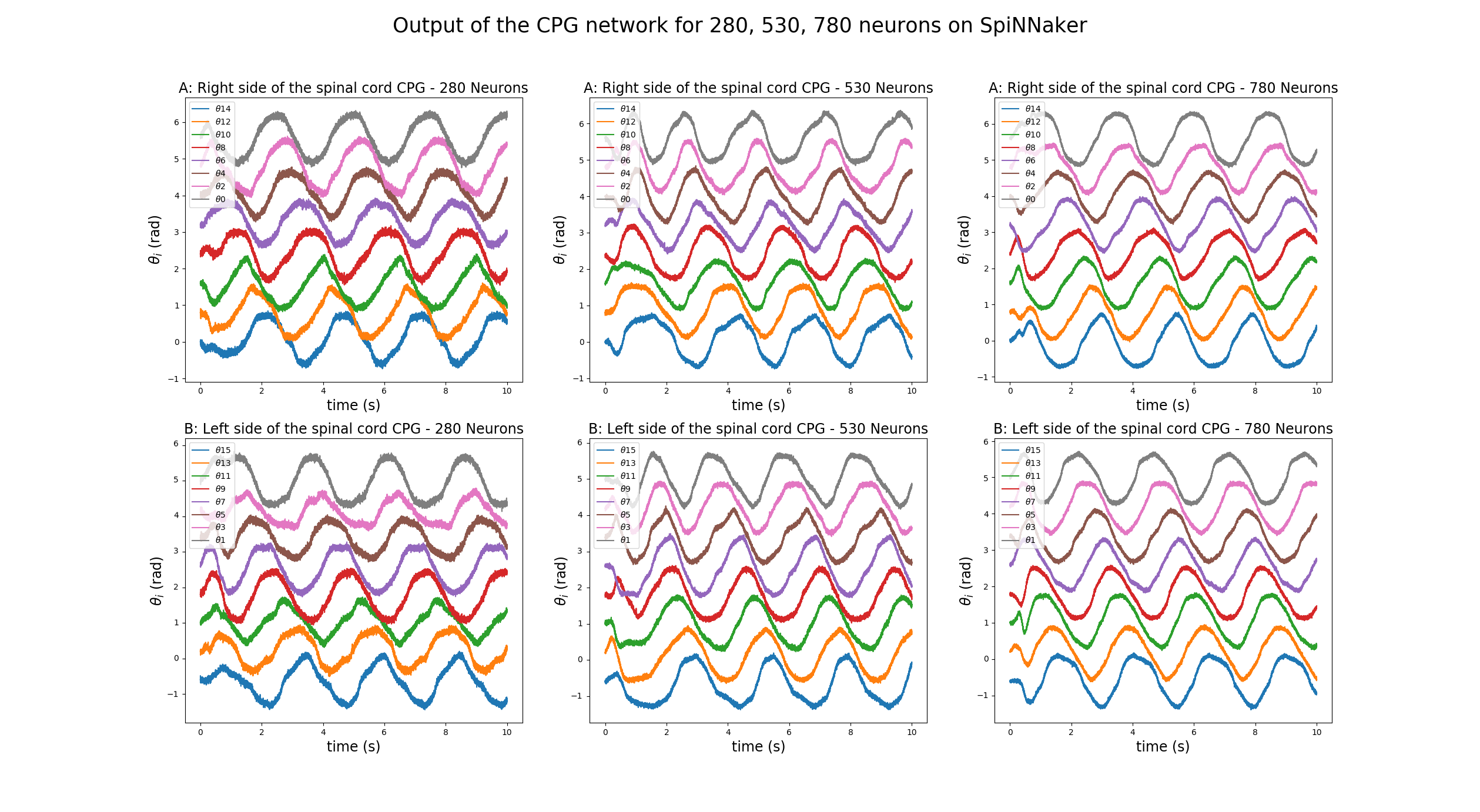}
\caption{The output of the network for different number of neurons per oscillatory population when executed on SpiNNaker. On SpiNNaker the output of the network is relatively accurate and better than the CPU even for a small number of neurons. The weights were trained with a random seed of 0. Note that high-frequency filtering is applied by default on the output signals.}
\label{fig:different_neurons_spinnaker}
\end{figure}

\begin{figure}[ht!]
\centering
\includegraphics[width=1\textwidth]{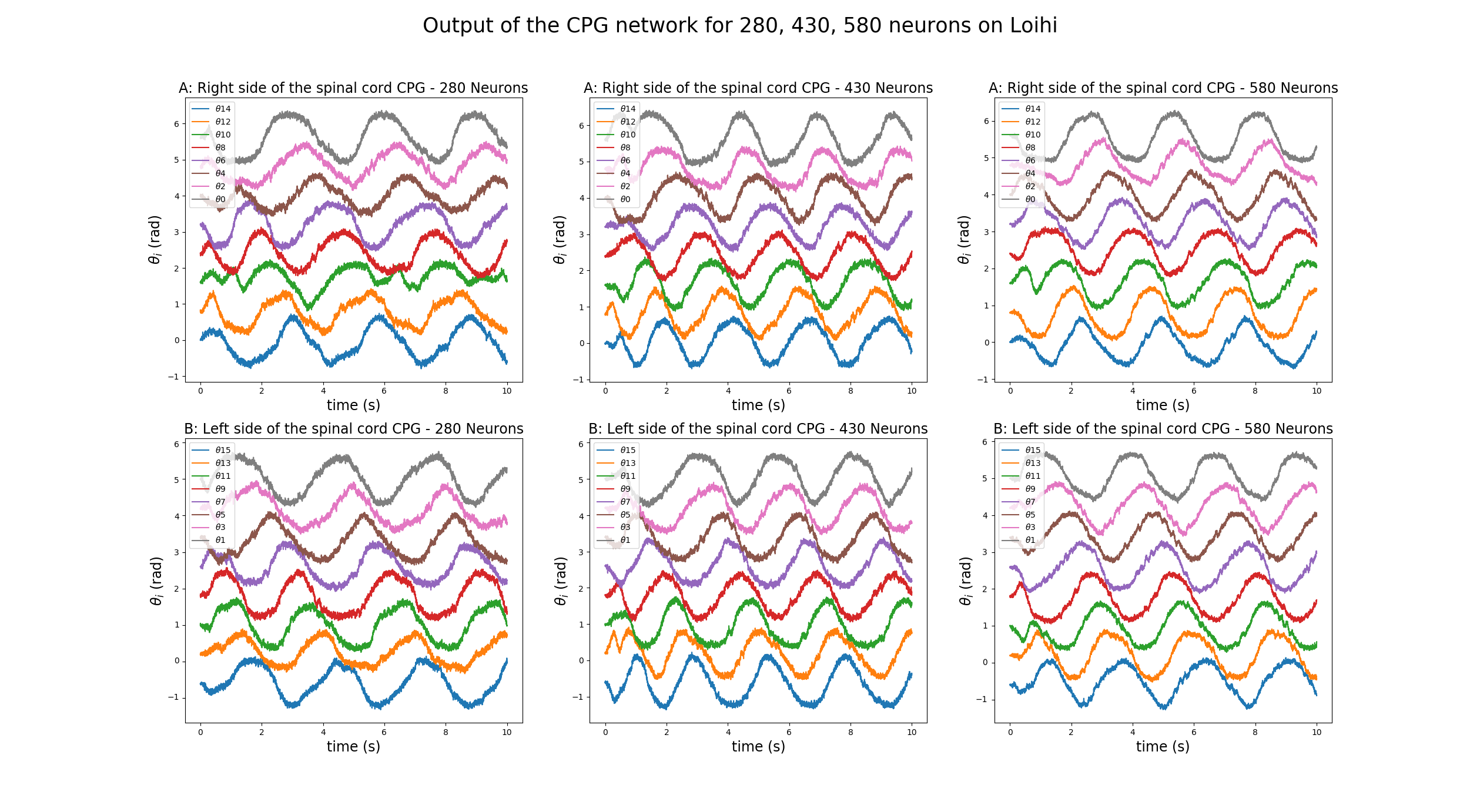}
\caption{The output of the network for different number of neurons per oscillatory population when executed on Loihi. The results have similar accuracy as SpiNNaker and perform better than the CPU for a low number of neurons. The weights were trained using the random seed 0. Note that high-frequency filtering is applied by default on the output signals.}
\label{fig:different_neurons_loihi}
\end{figure}

For robotics applications it is important that the control signals are generated in real-time. In order to be able to control a robot with the two neuromorphic boards that we examined, the quality of the generated signals has to be similar to the one coming from the CPU. Such comparison of the quality for a simulation of 10 secs can be seen in Figures \ref{fig:different_neurons_loihi} and \ref{fig:different_neurons_spinnaker}. As can be observed, the signals are of better quality than the CPU for a low number of neurons. The quality of the produced signals depends heavily on the number of neurons that are used to represent them. 
Due to limitations arising from the architecture of the two neuromorphic boards we tested, the total number of neurons that we could run on a SpiNNaker board is limited to 30000, for a Loihi board the limitations are reached at a similar number of neurons when no probes for measuring the networks output are used. With probes the limit on Loihi is reached at approximately 22000 neurons. The concept of a probe corresponds to a software construct that can be used to collect simulation data from the neuron activity, energy consumption etc. They are used to record the decoded output value of the neural population representing the oscillatory centres.

\begin{figure}[ht!]
\centering
\includegraphics[width=1\textwidth]{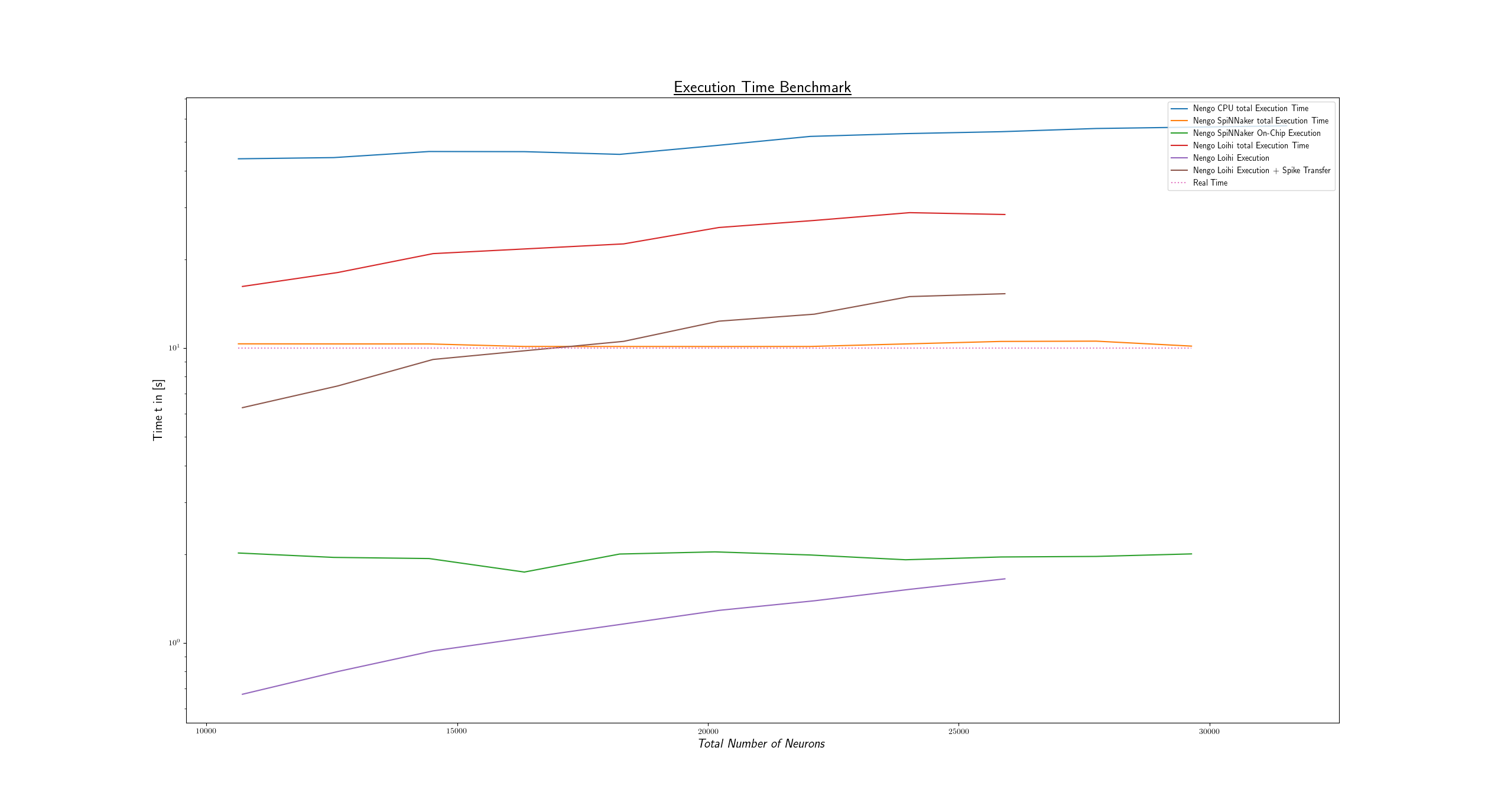}
\caption{Runtime of a 10 seconds experiment for various number of neurons per platform. The total execution time in SpiNNaker is referring to the complete execution cycle from the moment the simulation is launched to the moment the execution data is collected, likewise in Loihi. It is important to note that these values represent the execution of Nengo on the neuromorphic hardware from the perspective of an application developer, treating the hardware as a black box. The SpiNNaker on-chip execution time measures only the time spent on the board for the execution of the network. The Loihi execution measures the execution time reported by Loihi and represents the actual time spent executing the network. The execution + spike transfer represents the execution time plus the time spent during the exchange of spikes between the Loihi board and the CPU. The reasoning behind these benchmarks is to demonstrate that the times spent on the chip are very low compared to real-time and the rest of the times is spent on IO operations or other operations induced by the software. For a more detailed breakdown of the execution times in Loihi see also Figure \ref{fig:loihi_breakdown}. It can be observed that the actual execution time on the boards is much faster than real-time, showing that neuromorphic hardware is a great candidate for running the CPG model in real-time.}
\label{fig:runtime}
\end{figure}

\begin{figure}[ht!]
\centering
\includegraphics[width=1\textwidth]{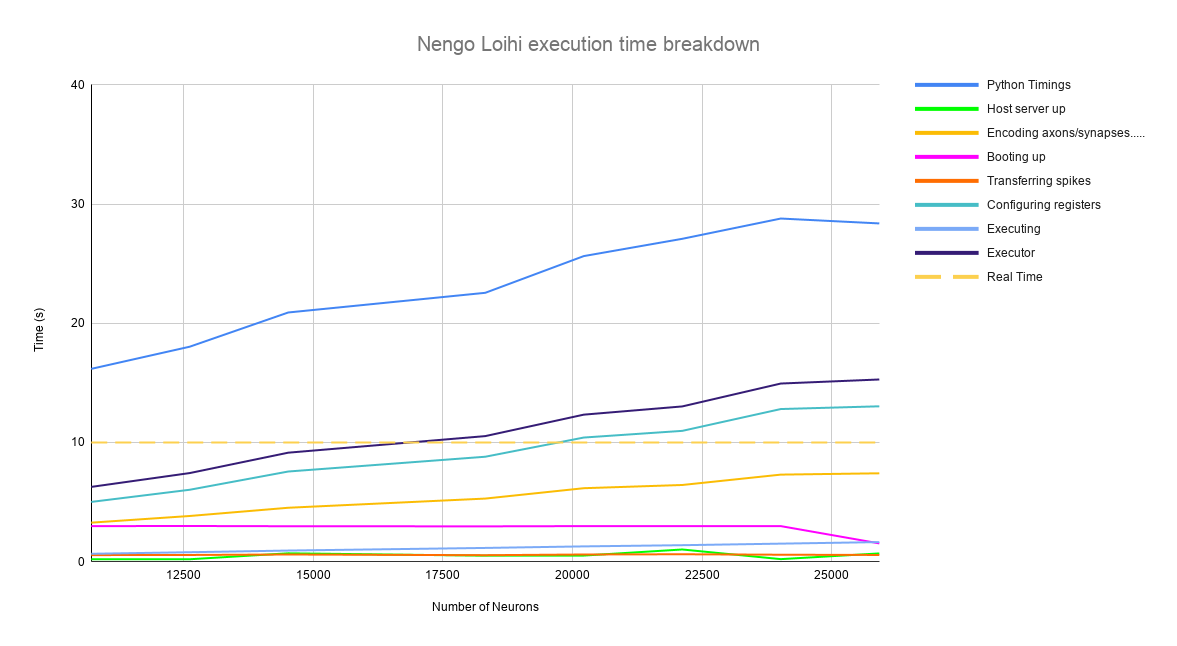}
\caption{Breakdown of total execution time on the Loihi chip into different parts for 10 seconds of simulation time and increasing neurons. Python timings refer to the execution of the network from an application developer's point of view and include all the software and IO induced times. The Executing series shows the actual execution time on the chip and is linearly increasing as the number of neurons increase. The Executor series includes both the execution and the transferring of spikes between the board and the CPU. It should be noted that these two processes can be performed in parallel. The times spent during the setup and initialization phases (Host server up, encoding axons/synapses, booting the board, configuring registers) are performed only once and their relative duration is less significant if the simulation time increases, see also \ref{fig:loihi_larger_simulation_times}}
\label{fig:loihi_breakdown}
\end{figure}

A more detailed comparison of the runtime performance for the different platforms can be see in figure \ref{fig:runtime}. What we observed during the execution on the neuromorphic chips is that most of the time is spent during phases other than the network execution, mostly during the initialization phase where the network configuration is being setup, and during input-output(I/O) operations such as the transfer of spikes between the neuromorphic board and the host computer. This is especially true for the Loihi board, as can be observed in figure \ref{fig:loihi_breakdown}, where the actual execution of the network is around 1 second for 10 seconds of simulation time, almost 10 times faster than real-time, slightly increasing as the network's size increases. In contrast, most of the time during execution is spent on other operations such as the exchange of spikes. It is clear, that this is the main bottleneck of Loihi's execution time. 
SpiNNaker on the other hand, and especially the execution of spiking networks on SpiNNaker through Nengo, is already optimized for real-time execution. This is the reason why the total operation of SpiNNaker including I/O operations and network execution is staying almost real-time. It should be noted that this time also includes waiting times induced by Nengo to make sure the simulation runs in real-time. The network itself is executed on SpiNNaker at around 2 seconds, marking a slightly slower execution time than Loihi.

\begin{figure}[ht!]
\centering
\includegraphics[width=1\textwidth]{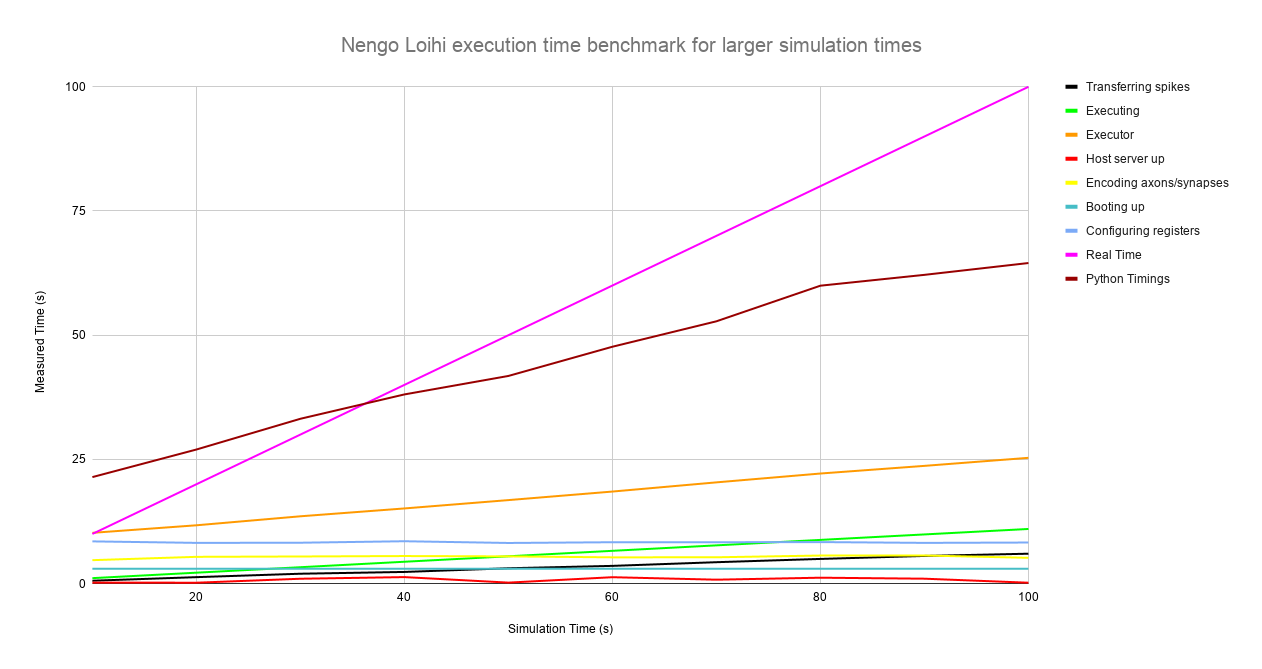}
\caption{Nengo Loihi execution times when the simulation time increases. All the benchmarks were performed with a network with 450 neurons per oscillatory center. In this figure it is evident that the initialization and setup times play an increasingly less significant role as the simulation time increases, making it possible to execute the network in real-time after roughly 35 secs of simulation time. This is important from the perspective of the application developer as it is taking into account all the software and IO bottlenecks, which usually treats the chips as black boxes and optimizes on the software and network layer. From the figure we can observe that the times spent during the operation of the chip are on the transfer of spikes and on the actual execution, which increase linearly in time, whereas all the other times remain relatively stable.}
\label{fig:loihi_larger_simulation_times}
\end{figure}

A more detailed analysis of the time spent during the execution of the network on Loihi during larger simulation times is provided in figure \ref{fig:loihi_larger_simulation_times}. To explain the observations it is useful to separate the operation of the board in three distinct phases. The first would be the initialization and setup phase which includes software overhead, overhead to boot the board, setup of the host server, compilation of neurons and synapses on the board and which is performed only once. The second phase would be the loading of the spikes into the neuromorphic board which can be done in parallel with the execution of the network, or before the execution of the simulation. The third phase corresponds to the actual execution on the board. From these findings we can conclude that as soon as the execution of the network is separated from the setup it can perform much faster than real-time. It should be noted that these metrics are relevant for this specific neural network and do not provide an accurate metric for other types of models. 

\begin{figure}[ht!]
\centering
\includegraphics[width=1\textwidth]{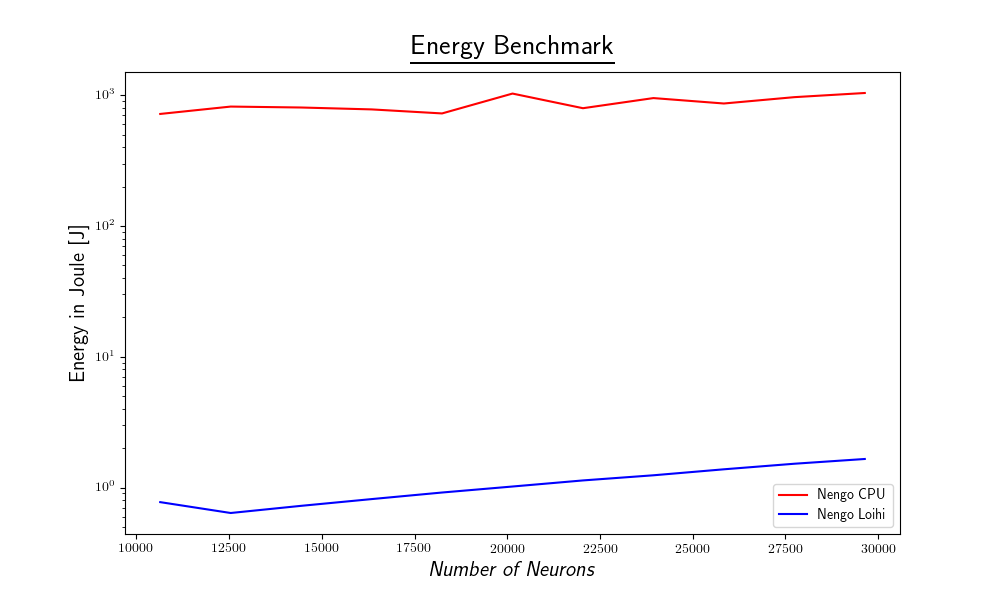}
\caption{Energy Benchmark of the CPG with Nengo Loihi and Nengo CPU, measured with built-in energy probes in Loihi and with the RAPL interface on the CPU. Is it clear that the energy consumption on the chip is orders of magnitude smaller that the consumption on the CPU. }
\label{fig:energy_benchmark}
\end{figure}

Due to software limitations it was not possible to provide accurate energy benchmarks for the SpiNNaker board. However, a comparison of the energy consumption between a CPU and Loihi is provided in figure \ref{fig:energy_benchmark}. On Loihi the energy consumption was measured with the built in time and energy probes. For measuring the energy consumption on the CPU, the RAPL interface was used. RAPL is an Intel processor feature that provides the ability of monitoring and controlling the SoC power consumption \cite{Intel64IA32}. As the power measurement control domain we used the \textsc{package} domain which includes the energy consumption of all cores, integrated graphics and other uncore components like caches and memory controllers. For the actual measurement, a framework developed by \cite{pereira_energy_2017} was used.

As a result, in figure \ref{fig:energy_benchmark} you can see that the energy consumption of the Loihi chip is by three orders of magnitude lower than executing the same network with Nengo CPU. This shows neuromorphic hardware can deliver significant energy reductions for executing spiking neural networks when compared to traditional CPU architectures. 
 
\section{Conclusions} \label{section:conclusion}
In this paper we presented a Spiking Central Pattern Generator based on a high-level system of abstract coupled Hopf-like oscillators that can run on both software and neuromorphic hardware. The method which we used can be generalized to any type of similar CPG controller. Our model is highly parametrizable, and is an excellent candidate for optimization methods. With different parametrizations it can provide a vast number of possible synchronized gaits, f.e. travelling and standing waves. Our method enables us to smoothly control a lamprey robot that with regulation of the high-level drive adapts to various simulation scenarios. We presented a closed-loop neurorobotics simulation within the Neurorobotics Platform achieving multiple locomotor tasks. Lastly, we showed that running the controller on neuromorphic hardware can achieve real-time operation and has potential advantages in terms of energy efficiency and computational speed.

Our work is related to other works in the field that attempt to provide insight on the performance of neuromorphic hardware. In particular, SpiNNaker was benchmarked for its performance in terms of energy efficiency and computational speed with similar accuracy, to an HPC system running a full-scale microcircuit of the human cortex model \cite{van_albada_performance_2018}. It was shown that for such complex models the energy consumption per synaptic event, which provides an estimate of the energy efficiency is \SI{5.9}{\micro J}, close to the \SI{5.8}{\micro J} consumed by the HPC system. However for simpler models, closer in terms of synaptic connections and number of neurons to the model that we employ, the cost per synaptic event can be as low as \SI{8}{\nano J} \cite{stromatias_power_2013}. Similarly, in \cite{Akopyan2015} they compared the performance of an IBM TrueNorth neuromorphic chip running a set of computer vision neural networks with the performance of a dual 2.4 GHz E5-2440 processor x86 system, as well as a Blue Gene/Q system with up to 32 compute cards and found two to three orders of execution time speedup and five orders of magnitude less energy consumption compared to the non-neuromorphic systems. Blouw et al. \cite{blouw_benchmarking_2019} showed that the energy performance of Intel's Loihi chip compared to the Movidius Neural Compute Stick, Nvidia’s Jetson TX1, a CPU, and a GPU was significantly lower (5.3x, 20.5x, 23.2x, 109.1x times respectively), for a keyword spotting task. However it should be noted that generating precise energy consumption benchmarks is a cumbersome task, and often the claims about the theoretical energy efficiency of neuromorphic hardware are not accompanied with the corresponding metrics.

\subsection{Future work}\label{section:future}
In order to study the challenges presented in animal swimming locomotion, a realistic simulation framework that can model all the different aspects of the physical world is necessary. The dynamics of the system, the control part, and their communication and synchronization is already solved in the Neurorobotics Platform, but a realistic fluid simulation is still missing. We are planning to address this problem and present a unified framework in our future works. This would allow providing realistic force feedback in the control loop, thus enabling the generation of more complex computational models.

Furthermore, our CPG model can be enriched with various form of environmental or sensory feedback, which can be incorporated into the model itself. Sensory data such as stretch receptors, high-level cognitive controllers that regulate the tonic drive are examples of this type of feedback.

One natural continuation of our work would be the transfer of the control framework on a real robot, such as the Amphibot. This is currently limited by the size of the SpiNNaker board that would prevent it from being fitted on the robot. However Loihi comes with a USB stick that is more compact in size and would potentially fit on the robot. One important consideration would be waterproofing the neuromorphic boards, as well as making sure that the changes induced in the dynamics of the system by the extra weight would be negligible. 

\section*{Acknowledgments}
The authors would like to thank Peter Blouw and Eric Hunsberger from Applied Brain Research for their valuable help on setting up the Nengo simulations and David Florey,Yulia Sandamirskaya and Andreas Wild from Intel for their help with the Loihi simulation and interpretation of results.

\section{Figures}

\clearpage
\bibliographystyle{unsrt}  
\bibliography{references}  %%% Remove comment to use the external .bib file (using bibtex).

\begin{thebibliography}{10}

\bibitem{knoll_neurorobotics2016}
A.~Knoll and Marc-Oliver Gewaltig.
\newblock Neurorobotics : A strategic pillar of the human brain project.
\newblock 2016.

\bibitem{capolei_biomimetic_2019}
Marie~Claire Capolei, Emmanouil Angelidis, Egidio Falotico, Henrik Lund, and
  Silvia Tolu.
\newblock A {Biomimetic} {Control} {Method} {Increases} the {Adaptability} of a
  {Humanoid} {Robot} {Acting} in a {Dynamic} {Environment}.
\newblock {\em Frontiers in Neurorobotics}, 13, August 2019.

\bibitem{garrido_alcazar_distributed_2013}
Jesus~A. Garrido~Alcazar, Niceto~Rafael Luque, Egidio D‘Angelo, and Eduardo
  Ros.
\newblock Distributed cerebellar plasticity implements adaptable gain control
  in a manipulation task: a closed-loop robotic simulation.
\newblock {\em Frontiers in Neural Circuits}, 7, 2013.

\bibitem{kaiser2019embodied2}
Jacques Kaiser, Alexander Friedrich, J.~Camilo~Vasquez Tieck, Daniel Reichard,
  Arne Roennau, Emre Neftci, and Rüdiger Dillmann.
\newblock Embodied neuromorphic vision with event-driven random
  backpropagation, 2019.

\bibitem{kaiser2019embodied}
Jacques Kaiser, Michael Hoff, Andreas Konle, Juan~Camilo Vasquez~Tieck, David
  Kappel, Daniel Reichard, Anand Subramoney, Robert Legenstein, Arne Roennau,
  Wolfgang Maass, et~al.
\newblock Embodied synaptic plasticity with online reinforcement learning.
\newblock {\em Frontiers in Neurorobotics}, 13:81, 2019.

\bibitem{bornet_running_2019}
Alban Bornet, Jacques Kaiser, Alexander Kroner, Egidio Falotico, Alessandro
  Ambrosano, Kepa Cantero, Michael~H. Herzog, and Gregory Francis.
\newblock Running {Large}-{Scale} {Simulations} on the {Neurorobotics}
  {Platform} to {Understand} {Vision} – {The} {Case} of {Visual} {Crowding}.
\newblock {\em Frontiers in Neurorobotics}, 13, 2019.
\newblock Publisher: Frontiers.

\bibitem{ijspeert_swimming_2007}
A.~J. Ijspeert, A.~Crespi, D.~Ryczko, and J.-M. Cabelguen.
\newblock From {Swimming} to {Walking} with a {Salamander} {Robot} {Driven} by
  a {Spinal} {Cord} {Model}.
\newblock {\em Science}, 315(5817):1416--1420, March 2007.

\bibitem{Bing_2017}
Zhenshan Bing, Long Cheng, Guang Chen, Florian Röhrbein, Kai Huang, and Alois
  Knoll.
\newblock Towards autonomous locomotion: {CPG}-based control of smooth 3d
  slithering gait transition of a snake-like robot.
\newblock {\em Bioinspiration {\&} Biomimetics}, 12(3):035001, apr 2017.

\bibitem{prescott_robot_2006}
Tony~J. Prescott, Fernando~M. Montes~González, Kevin Gurney, Mark~D.
  Humphries, and Peter Redgrave.
\newblock A robot model of the basal ganglia: behavior and intrinsic
  processing.
\newblock {\em Neural Networks: The Official Journal of the International
  Neural Network Society}, 19(1):31--61, January 2006.

\bibitem{maass_networks_1997}
Wolfgang Maass.
\newblock Networks of spiking neurons: {The} third generation of neural network
  models.
\newblock {\em Neural Networks}, 10(9):1659--1671, December 1997.

\bibitem{Davies_2018}
M.~{Davies}, N.~{Srinivasa}, T.~{Lin}, G.~{Chinya}, Y.~{Cao}, S.~H. {Choday},
  G.~{Dimou}, P.~{Joshi}, N.~{Imam}, S.~{Jain}, Y.~{Liao}, C.~{Lin},
  A.~{Lines}, R.~{Liu}, D.~{Mathaikutty}, S.~{McCoy}, A.~{Paul}, J.~{Tse},
  G.~{Venkataramanan}, Y.~{Weng}, A.~{Wild}, Y.~{Yang}, and H.~{Wang}.
\newblock Loihi: A neuromorphic manycore processor with on-chip learning.
\newblock {\em IEEE Micro}, 38(1):82--99, January 2018.

\bibitem{Akopyan2015}
F.~{Akopyan}, J.~{Sawada}, A.~{Cassidy}, R.~{Alvarez-Icaza}, J.~{Arthur},
  P.~{Merolla}, N.~{Imam}, Y.~{Nakamura}, P.~{Datta}, G.~{Nam}, B.~{Taba},
  M.~{Beakes}, B.~{Brezzo}, J.~B. {Kuang}, R.~{Manohar}, W.~P. {Risk},
  B.~{Jackson}, and D.~S. {Modha}.
\newblock Truenorth: Design and tool flow of a 65 mw 1 million neuron
  programmable neurosynaptic chip.
\newblock {\em IEEE Transactions on Computer-Aided Design of Integrated
  Circuits and Systems}, 34(10):1537--1557, Oct 2015.

\bibitem{SpiNNaker}
Steve~B. Furber, Francesco Galluppi, Steve Temple, and Luis~A. Plana.
\newblock The spinnaker project.
\newblock {\em IEEE. Proceedings}, 102(5):652--665, 2014.

\bibitem{Schemmel_2010}
J.~{Schemmel}, D.~{Briiderle}, A.~{Griibl}, M.~{Hock}, K.~{Meier}, and
  S.~{Millner}.
\newblock A wafer-scale neuromorphic hardware system for large-scale neural
  modeling.
\newblock In {\em Proceedings of 2010 IEEE International Symposium on Circuits
  and Systems}, pages 1947--1950, May 2010.

\bibitem{arena_cpg}
Paolo Arena.
\newblock The central pattern generator: A paradigm for artificial locomotion.
\newblock {\em Soft Computing}, 4:251--266, 01 2000.

\bibitem{ijspeert_central_2008}
Auke~Jan Ijspeert.
\newblock Central pattern generators for locomotion control in animals and
  robots: {A} review.
\newblock {\em Neural Networks}, 21(4):642--653, May 2008.

\bibitem{grillner_biological_2006}
Sten Grillner.
\newblock Biological pattern generation: the cellular and computational logic
  of networks in motion.
\newblock {\em Neuron}, 52(5):751--766, December 2006.

\bibitem{grillner_motor_2003}
Sten Grillner.
\newblock The motor infrastructure: from ion channels to neuronal networks.
\newblock {\em Nature Reviews Neuroscience}, 4(7):573--586, July 2003.

\bibitem{survey_cpg_yu}
Junzhi Yu, M.~Tan, Jian Chen, and Jianwei Zhang.
\newblock A survey on cpg-inspired control models and system implementation.
\newblock {\em IEEE transactions on neural networks and learning systems},
  25:441--56, 03 2014.

\bibitem{crespi_amphibot_nodate}
A.~Crespi and A.J. Ijspeert.
\newblock Amphibot ii: An amphibious snake robot that crawls and swims using a
  central pattern generator.
\newblock {\em Proceedings of the 9th International Conference on Climbing and
  Walking Robots (CLAWAR 2006)}, 01 2006.

\bibitem{inoue_neural_2004}
K.~Inoue, Shugen Ma, and Chenghua Jin.
\newblock Neural oscillator network-based controller for meandering locomotion
  of snake-like robots.
\newblock In {\em {IEEE} {International} {Conference} on {Robotics} and
  {Automation}, 2004. {Proceedings}. {ICRA} '04. 2004}, volume~5, pages
  5064--5069, April 2004.
\newblock ISSN: 1050-4729.

\bibitem{donati_novel_2016}
Elisa Donati, Giacomo Indiveri, and Cesare Stefanini.
\newblock A novel spiking {CPG}-based implementation system to control a
  lamprey robot.
\newblock In {\em 2016 6th {IEEE} {International} {Conference} on {Biomedical}
  {Robotics} and {Biomechatronics} ({BioRob})}, pages 1364--1364, June 2016.
\newblock ISSN: 2155-1782.

\bibitem{wang_cpg-inspired_2017}
Zhelong Wang, Qin Gao, and Hongyu Zhao.
\newblock {CPG}-{Inspired} {Locomotion} {Control} for a {Snake} {Robot}
  {Basing} on {Nonlinear} {Oscillators}.
\newblock {\em Journal of Intelligent \& Robotic Systems}, 85(2):209--227,
  February 2017.

\bibitem{cuevas-arteaga_spinnaker_2017}
Brayan Cuevas-Arteaga, Juan~Pedro Dominguez-Morales, Horacio Rostro-Gonzalez,
  Andres Espinal, Angel Jiménez-Fernandez, Francisco Gómez-Rodríguez, and
  Alejandro Linares-Barranco.
\newblock A spinnaker application: Design, implementation and validation of
  scpgs.
\newblock volume 10305, 06 2017.

\bibitem{russel_scpg}
Alex Russell, Garrick Orchard, and Ralph Etienne-Cummings.
\newblock Configuring of spiking central pattern generator networks for bipedal
  walking using genetic algorthms.
\newblock pages 1525 -- 1528, 06 2007.

\bibitem{lewis_cpg_design}
M.~Lewis, Francesco Tenore, and Ralph Etienne-Cummings.
\newblock Cpg design using inhibitory networks.
\newblock volume 2005, pages 3682--3687, 01 2005.

\bibitem{bouganis_training_2010}
Alexandros Bouganis and Murray Shanahan.
\newblock Training a spiking neural network to control a 4-{DoF} robotic arm
  based on {Spike} {Timing}-{Dependent} {Plasticity}.
\newblock In {\em The 2010 {International} {Joint} {Conference} on {Neural}
  {Networks} ({IJCNN})}, pages 1--8, Barcelona, Spain, July 2010. IEEE.

\bibitem{menon_controlling_2014}
Samir Menon, Sam Fok, Alex Neckar, Oussama Khatib, and Kwabena Boahen.
\newblock Controlling articulated robots in task-space with spiking silicon
  neurons.
\newblock In {\em 5th {IEEE} {RAS}/{EMBS} {International} {Conference} on
  {Biomedical} {Robotics} and {Biomechatronics}}, pages 181--186, Sao Paulo,
  Brazil, August 2014. IEEE.

\bibitem{espinal_design_2016}
Andres Espinal, Horacio Rostro-Gonzalez, Martin Carpio, Erick~I.
  Guerra-Hernandez, Manuel Ornelas-Rodriguez, and Marco Sotelo-Figueroa.
\newblock Design of {Spiking} {Central} {Pattern} {Generators} for {Multiple}
  {Locomotion} {Gaits} in {Hexapod} {Robots} by {Christiansen} {Grammar}
  {Evolution}.
\newblock {\em Frontiers in Neurorobotics}, 10, 2016.

\bibitem{gutierrez-galan_neuropod:_2019}
Daniel Gutierrez-Galan, Juan~Pedro Dominguez-Morales, Fernando Perez-Pena, and
  Alejandro Linares-Barranco.
\newblock {NeuroPod}: a real-time neuromorphic spiking {CPG} applied to
  robotics.
\newblock {\em Neurocomputing}, page S0925231219315644, November 2019.
\newblock arXiv: 1904.11243.

\bibitem{eliasmith_rethinking_2000}
Chris Eliasmith and Charles~H. Anderson.
\newblock Rethinking central pattern generators: {A} general approach.
\newblock {\em Neurocomputing}, 32-33:735--740, June 2000.

\bibitem{crespi_amphibot_2005}
Alessandro Crespi, André Badertscher, André Guignard, and Auke~Jan Ijspeert.
\newblock {AmphiBot} {I}: an amphibious snake-like robot.
\newblock {\em Robotics and Autonomous Systems}, 50(4):163--175, March 2005.

\bibitem{bicanski_decoding_2013}
Andrej Bicanski, Dimitri Ryczko, Jérémie Knuesel, Nalin Harischandra, Vanessa
  Charrier, Örjan Ekeberg, Jean-Marie Cabelguen, and Auke~Jan Ijspeert.
\newblock Decoding the mechanisms of gait generation in salamanders by
  combining neurobiology, modeling and robotics.
\newblock {\em Biological Cybernetics}, 107(5):545--564, October 2013.

\bibitem{Gerstner:2002:SNM:583784}
Wulfram Gerstner and Werner Kistler.
\newblock {\em Spiking Neuron Models: An Introduction}.
\newblock Cambridge University Press, New York, NY, USA, 2002.

\bibitem{markram_spike-timing-dependent_2012}
H.~Markram, W.~Gerstner, and P.~J. Sjöström.
\newblock Spike-{Timing}-{Dependent} {Plasticity}: {A} {Comprehensive}
  {Overview}.
\newblock {\em Frontiers in Synaptic Neuroscience}, 4, July 2012.

\bibitem{bellec_long_2018}
Guillaume Bellec, Darjan Salaj, Anand Subramoney, Robert Legenstein, and
  Wolfgang Maass.
\newblock Long short-term memory and {Learning}-to-learn in networks of spiking
  neurons.
\newblock In S.~Bengio, H.~Wallach, H.~Larochelle, K.~Grauman, N.~Cesa-Bianchi,
  and R.~Garnett, editors, {\em Advances in {Neural} {Information} {Processing}
  {Systems} 31}, pages 787--797. Curran Associates, Inc., 2018.

\bibitem{bellec_biologically_2019}
Guillaume Bellec, Franz Scherr, Elias Hajek, Darjan Salaj, Robert Legenstein,
  and Wolfgang Maass.
\newblock Biologically inspired alternatives to backpropagation through time
  for learning in recurrent neural nets.
\newblock {\em arXiv:1901.09049 [cs]}, February 2019.
\newblock arXiv: 1901.09049.

\bibitem{bekolay_nengo:_2014}
Trevor Bekolay, James Bergstra, Eric Hunsberger, Travis DeWolf, Terrence~C.
  Stewart, Daniel Rasmussen, Xuan Choo, Aaron~Russell Voelker, and Chris
  Eliasmith.
\newblock Nengo: a {Python} tool for building large-scale functional brain
  models.
\newblock {\em Frontiers in Neuroinformatics}, 7, 2014.

\bibitem{eliasmith_neural_2003}
Chris Eliasmith and Charles Anderson.
\newblock Neural engineering: Computation, representation, and dynamics in
  neurobiological systems.
\newblock {\em IEEE Transactions on Neural Networks}, 15(2):528--529, March
  2004.

\bibitem{Stewart2012ATO}
Terrence~C. Stewart.
\newblock A technical overview of the neural engineering framework.
\newblock Technical report, Centre for Theoretical Neuroscience, 2012.

\bibitem{falotico_connecting_2017}
Egidio Falotico, Lorenzo Vannucci, Alessandro Ambrosano, Ugo Albanese, Stefan
  Ulbrich, Juan~Camilo Vasquez~Tieck, Georg Hinkel, Jacques Kaiser, Igor Peric,
  Oliver Denninger, Nino Cauli, Murat Kirtay, Arne Roennau, Gudrun Klinker,
  Axel Von~Arnim, Luc Guyot, Daniel Peppicelli, Pablo Martínez-Cañada,
  Eduardo Ros, Patrick Maier, Sandro Weber, Manuel Huber, David Plecher,
  Florian Röhrbein, Stefan Deser, Alina Roitberg, Patrick van~der Smagt,
  Rüdiger Dillman, Paul Levi, Cecilia Laschi, Alois~C. Knoll, and Marc-Oliver
  Gewaltig.
\newblock Connecting {Artificial} {Brains} to {Robots} in a {Comprehensive}
  {Simulation} {Framework}: {The} {Neurorobotics} {Platform}.
\newblock {\em Frontiers in Neurorobotics}, 11, 2017.

\bibitem{Quigley09}
Morgan Quigley, Brian Gerkey, Ken Conley, Josh Faust, Tully Foote, Jeremy
  Leibs, Eric Berger, Rob Wheeler, and Andrew Ng.
\newblock Ros: an open-source robot operating system.
\newblock In {\em Proc. of the IEEE Intl. Conf. on Robotics and Automation
  (ICRA) Workshop on Open Source Robotics}, Kobe, Japan, May 2009.

\bibitem{Koenig:gazebo}
N.~{Koenig} and A.~{Howard}.
\newblock Design and use paradigms for gazebo, an open-source multi-robot
  simulator.
\newblock In {\em 2004 IEEE/RSJ International Conference on Intelligent Robots
  and Systems (IROS) (IEEE Cat. No.04CH37566)}, volume~3, pages 2149--2154
  vol.3, 2004.

\bibitem{Gewaltig:NEST}
Marc-Oliver Gewaltig and Markus Diesmann.
\newblock Nest (neural simulation tool).
\newblock {\em Scholarpedia}, 2(4):1430, 2007.

\bibitem{ekeberg_combined_1993}
Örjan Ekeberg.
\newblock A combined neuronal and mechanical model of fish swimming.
\newblock {\em Biological Cybernetics}, 69(5):363--374, October 1993.

\bibitem{an_roadmap_2018}
Hongyu An, Kangjun Bai, and Yang Yi.
\newblock {\em The Roadmap to Realizing Memristive Three-dimensional
  Neuromorphic Computing System}.
\newblock November 2018.

\bibitem{Intel64IA32}
{{Intel}}® 64 and {{IA}}-32 {{Architectures Software Developer}}’s
  {{Manual}}, {{Volume 3B}}: {{System Programming Guide}}, {{Part}} 2.
\newblock page 582.

\bibitem{pereira_energy_2017}
Rui Pereira, Marco Couto, Francisco Ribeiro, Rui Rua, Jácome Cunha,
  João~Paulo Fernandes, and João Saraiva.
\newblock Energy efficiency across programming languages: how do energy, time,
  and memory relate?
\newblock In {\em Proceedings of the 10th {ACM} {SIGPLAN} {International}
  {Conference} on {Software} {Language} {Engineering}}, pages 256--267,
  Vancouver BC Canada, October 2017. ACM.

\bibitem{van_albada_performance_2018}
Sacha~J. van Albada, Andrew~G. Rowley, Johanna Senk, Michael Hopkins,
  Maximilian Schmidt, Alan~B. Stokes, David~R. Lester, Markus Diesmann, and
  Steve~B. Furber.
\newblock Performance {Comparison} of the {Digital} {Neuromorphic} {Hardware}
  {SpiNNaker} and the {Neural} {Network} {Simulation} {Software} {NEST} for a
  {Full}-{Scale} {Cortical} {Microcircuit} {Model}.
\newblock {\em Frontiers in Neuroscience}, 12, 2018.
\newblock Publisher: Frontiers.

\bibitem{stromatias_power_2013}
Evangelos Stromatias, Francesco Galluppi, Cameron Patterson, and Steve Furber.
\newblock Power analysis of large-scale, real-time neural networks on
  {SpiNNaker}.
\newblock In {\em The 2013 {International} {Joint} {Conference} on {Neural}
  {Networks} ({IJCNN})}, pages 1--8, August 2013.
\newblock ISSN: 2161-4407.

\bibitem{blouw_benchmarking_2019}
Peter Blouw, Xuan Choo, Eric Hunsberger, and C.~Eliasmith.
\newblock Benchmarking {Keyword} {Spotting} {Efficiency} on {Neuromorphic}
  {Hardware}.
\newblock {\em NICE '19}, 2019.

\end{thebibliography}
%%% and comment out the ``thebibliography'' section.

%%% Comment out this section when you \bibliography{references} is enabled.
\end{document}